# 卷积神经网络在图像超分辨上的应用

田春伟 [1,2]，宋明键 [3]，左旺孟 [1]，杜博 [4]，张艳宁 [2,5]，张师超 [6]

（1. 哈尔滨工业大学 计算学部，黑龙江 哈尔滨 150001；2. 空天地海一体化大数据应用技术国家工程实验室，陕西 西安 710192；3. 西北工业大学 软件学院，陕西 西安 710072；4. 武汉大学 计算机学院，湖北 武汉 430072；5. 西北工业大学 计算机学院，陕西 西安 710072； 6. 广西师范大学 计算机科学与工程学院，广西 桂林 541004）

**摘　要**：卷积神经网络因强大的学习能力，已成为解决图像超分辨问题的主流方法。然而，用于解决图像超分辨的不同类型深度学习方法存在巨大的差异。目前，仅有少量文献能根据不同缩放方法来总结不同深度学习技术在图像超分辨上的区别和联系。因此，根据设备的负载能力和执行速度等介绍面向图像超分辨方法的卷积神经网络尤为重要。本文首先介绍面向图像超分辨的卷积神经网络基础，随后通过介绍基于双三次插值、最近邻插值、双线性插值、转置卷积、亚像素层、元上采样的卷积神经网络的图像超分辨方法，分析基于插值和模块化的卷积神经网络图像超分辨方法的区别与联系，并通过实验比较这些方法的性能。本文对潜在的研究点和挑战进行阐述并总结全文，旨在促进基于卷积神经网络的图像超分辨研究的发展。

# Application of convolutional neural networks in image super-resolution

TIAN Chunwei[1,2]，SONG Mingjian[3]，ZUO Wangmeng[1]，DU Bo[4]，ZHANG Yanning[2,5]，ZHANG Shichao[6]

(1. Faculty of Computing, Harbin Institute of Technology, Harbin 150001, China; 2. National Engineering Laboratory for Big Data Application Technology of Integrated Space, Air, Ground, and Sea, Xi'an 710192, China; 3. School of Software, Northwestern Polytechnical University, Xi'an 710072, China; 4. School of Computer Science, Wuhan University, Wuhan 430072, China; 5. School of Computer Science, Northwestern Polytechnical University, Xi'an 710072; 6. School of Computer Science and Engineering, Guangxi Normal University, Guilin 541004, China)

**Abstract**：Due to strong learning abilities of convolutional neural networks (CNNs), they have become mainstream methods for image super-resolution. However, there are big differences of different deep learning methods with different types. There is little literature to summarize relations and differences of different methods in image super-resolution. Thus, summarizing these literatures are important, according to loading capacity and execution speed of devices. This paper first introduces principles of CNNs in image super-resolution, then introduces CNNs based bicubic interpolation, nearest neighbor interpolation, bilinear interpolation, transposed convolution, sub-pixel layer, meta up-sampling for image super-resolution to analyze differences and relations of different CNNs based interpolations and modules, and compare performance of these methods by experiments. Finally, this paper gives potential research points and drawbacks and summarizes the whole paper, which can facilitate developments of CNNs in image super-resolution.

**收稿日期**：2024-09-19.
**基金项目**：国家自然科学基金项目（62201468）.

图像超分辨（super-resolution, SR）能通过机器学习方法将低分辨率（low-resolution, LR）图像重建成具有丰富细节信息和清晰纹理的高分辨率（high-resolution, HR）图像[1]。它通过提高图像质量来提高目标检测、图像分类等众多上层计算机视觉任务的性能，为卫星遥感、公共安防[2]和视频感知[3]等领域提供助力。

早期的图像超分辨方法如图像插值方法[4]、基于稀疏表示方法、基于局部嵌入方法、基于退化模型方法[5]等主要通过图像中像素之间的关系，提升图像的质量。其中，基于图像插值方法是简单和容易实现的，但无法很好地恢复高频信息和纹理特征[6]。为了解决此问题，基于稀疏表示方法利用图像集来训练字典，获得高分辨率图像[7]。虽然该方法在一定程度上能恢复图像细节信息，但仍会出现锯齿状问题[8]。为了获得更多图像细节信息，基于局部嵌入方法通过流形学习中局部线性嵌入 K 近邻算法来提高图像的像素质量。随后，学者通过高分辨（high resolution, HR）和低分辨率（low resolution, LR）图像间的退化关系进行建模，并通过优化目标函数与解函数重建 HR 图像[8]。例如，迭代反投影法[9]、凸集投影法[10]等方法均从退化模型的角度出发，利用投影获得更准确的参数、恢复更多图像细节和加快恢复高质量图像的速度[11-12]。为了提取更多的 LR 图像关键信息，优化退化关系，更多机器学习方法被用来学习对应的高频信息[13]。这些图像超分辨方法，结合有限的图像信息，获得高质量图像，不仅推动图像超分辨领域的发展[13]，也为智能交通和人脸识别[14]等计算机视觉应用提供有价值的方案。但是这些方法需要手动设置参数和优化方法等，影响图像超分辨任务的效率[15]。

为了解决这些问题，具有强大学习能力的深度学习技术被用来解决图像超分辨。例如：Dong 等[16]首次用端到端结构解决图像超分辨问题。具体为：它通过三层卷积结构把低分辨率图像以像素形式映射到高分辨率图像，该方法称为超分辨率卷积神经网络（super-resolution convolutional neural network, SRCNN）[16]。虽然 SRCNN 在图像超分辨性能上明显超过传统的图像超分辨方法，但 SRCNN 可扩展性差和特征维数过少，导致它没有被推广应用[17]。随后，Dong 等[18]提出快速的超分辨率卷积神经网络（fast super-resolution convolutional neural network, FSRCNN）来优化 SRCNN 模型，加快图像超分辨模型的训练效率。FSRCNN 主要通过深层的反卷积操作放大图像来代替图像预处理的双三次插值等操作，降低训练过程的效率。此外，FSRCNN 使用更小的卷积核和更多的映射层处理，改变特征维数并获得图像特征，提高图像质量。由于深度网络具有较深的特征提取能力和残差学习操作具有较强的浅层特征传递能力，残差卷积神经网络用来解决图像超分辨。例如：Kim 等[17]将残差学习操作用到卷积神经网络的末端构造深度超分辨率卷积神经网络（very deep super-resolution convolutional network, VDSR）提升预测图像的质量。此外递归网络通过递归连接卷积层不仅控制参数量，还能增强网络的表达能力，加快模型训练速度，提高图像重建效果[19-20]。为了增强网络的深度学习特征能力，密集连接融合网络的层次特征，增强模型的信号传播和特征重用能力，获得更多的图像细节信息，提升图像复原的质量[21-22]，Zhang 等[22]利用当前层与所有后层进行相连，增强深度网络的记忆能力，提高图像超分辨的性能。该方法不仅能防止深度网络的梯度消失或爆炸现象的出现，还能大大地提升获得图像的清晰度。为了提高图像超分辨的速度，注意力机制利用当前层引导之前层来获得显著性特征，加快图像重建效率。例如，Zhang 等[23]利用残差网络结构和通道注意力提取丰富的高频和低频信息，这不仅能解决深度网络难训练问题，还能增强图像超分辨的效果。Dai 等[24]利用二阶特征统计来自适应地重新缩放通道特征实现二阶注意力网络（second-order attention network, SAN），增强网络的判别能力，提高预测图像的清晰度。为了解决样本不足问题，生成对抗网络利用博弈思想设计生成器和判别器，实现图像超分辨模型。其中，生成器用于生成重建图像，判别器用于判断生成图像与真实图像的差距，通过二者的博弈来提高图像超分辨的性能。例如 Ledig 等[25]利用感知损失函数来训练生成对抗网络（super-resolution generative adversarial network, SRGAN），获得更多的纹理信息，增强预测图像的质量。为了解决伪影问题，Wang 等[26]利用没有批量标准化的残差密集块改进网络，并使用激活前的特征来改进感知损失提高恢复图像的质量。为了同时获得更优的视觉效果，Wang 等[27]组合感知损失和对抗损失来提高特征重构能力。

尽管卷积神经网络在图像超分辨上有很多研究[28]，但很少有相关文献系统地阐述图像超分辨领域中不同卷积神经网络的区别与联系。本文首先回

**通信作者**：张师超. E-mail：zhangsc@mailbox.gxnu.edu.cn.

顾了图像超分辨从传统方法到深度学习方法的发展历史，搜集整理了多个超分辨技术框架，分析总结了近百种不同角度、不同思路的超分辨方法，并通过系统性的对比研究，展示了它们的性能、优缺点、复杂性、挑战和潜在的研究要点等。具体为：本文首先介绍了图像超分辨技术近年来的发展状况，概述了从传统的图像超分辨方法到基于卷积神经网络的图像超分辨方法的演进过程。第二，本文从技术前身、基本原理与实际效果等角度，完整介绍了基于卷积神经网络图像超分辨率方法的经典网络框架。第三，根据不同的技术原理，本文将面向图像超分辨的卷积神经网络划分为两类，包括基于插值和基于模块化的卷积神经网络图像超分辨方法，并分析了用于非盲图像超分辨与盲超分辨的卷积神经网络的动机、实现、性能与差异。 第四，本文定性和定量分析这些算法在图像超分辨公共数据集上的性能。最后，本文指出了卷积神经网络在图像超分辨领域面临的挑战与未来发展方向。

## 1　面向图像超分辨的卷积神经网络的主流框架

尽管传统机器学习的图像处理方法可以利用先验知识提高图像超分辨的处理速度，但仍存在两点不足[29]：第一，机器学习方法需要手动设置参数；第二，需要通过复杂的优化算法来优化参数[30]。针对上述问题，基于深度学习的图像处理方法被提出。它依靠深度的网络结构，能自动学习图像特征，这不仅能避免手动设置参数和优化参数的使用，而且已广泛应用在图像分类、图像修复和图像超分辨率等图像处理任务中[31]。本文的总体结构如图 1 所示。

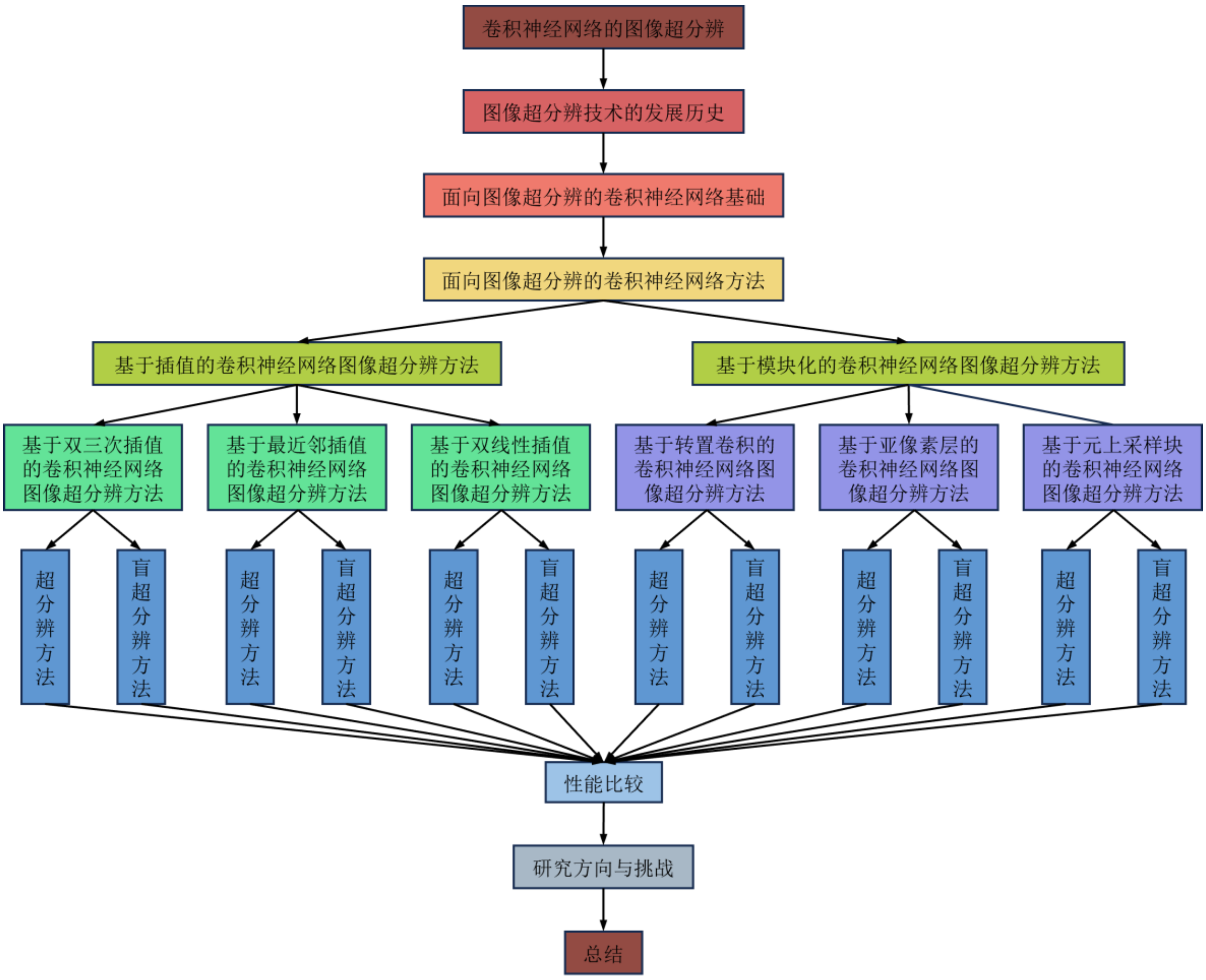

图 1　综述结构

**Fig.1 The Structure of the Review**

其中，本节通过介绍六种图像超分辨领域的经典的网络框架：超分辨率卷积神经网络

（super-resolution convolutional neural network, SRCNN）[16]、极深图像超分辨卷积网络（very deep convolutional networks for image super-resolution, VDSR）[17]、深度递归卷积网络（deeply-recursive convolutional network, DRCN）[20]、快速超分辨率卷积神经网络（fast super-resolution convolutional neural network, FSRCNN）[18]、超分辨率密集网络（super-resolution dense network, SRDenseNet）[21]和残差密集网络（residual dense network, RDN）[22]来了解基于深度学习的图像超分辨方面的研究，这是本篇针对卷积神经网络的图像超分辨技术进行讨论与研究的基础，具体如下：

为了提升图像超分辨的效果，Dong 等[16]利用三层卷积，提出了首个端到端的图像超分辨卷积神经网络 SRCNN，其网络结构如图 2 所示。该模型可以直接学习 LR 和 HR 之间的端到端映射，并将这种映射表示为以 LR 图像为输入，HR 图像为输出的深度卷积神经网络。与传统方法相比，该模型不仅能自动地学习参数，还使模型具有好的图像超分辨性能。该网络的具体实现如下：SRCNN 模型包括一个预处理操作和三个卷积层。首先，预处理操作对输入的 LR 图像进行双三次插值操作，通过上采样获得与目标相同大小的输入图像。其次，第一卷积层从 LR 图像中提取特征。第二卷积层将这些特征非线性地映射到 HR 的表示块中，获得高频特征[1]。最后，第三卷积层将获得高频特征图放在一个空间区域中，生成最终的 HR 图像。轻量化的结构、较高的执行效率、较强的鲁棒性，导致其实际效果优于传统的图像超分辨技术，这不仅是卷积神经网络在图像超分辨领域应用的基石，而且为该领域的不断发展提供了方向和框架[16]。

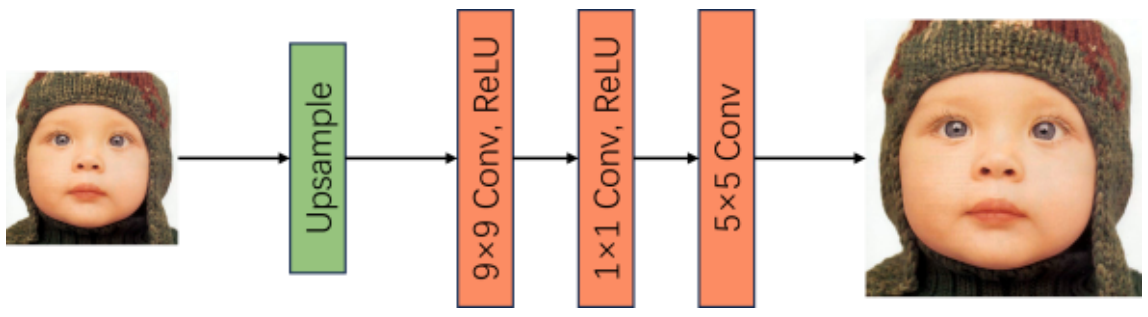

图 2 SRCNN 的网络结构

**Fig.2 The Network Structure of SRCNN**

SRCNN 虽然成功地将深度学习技术引入到图像超分辨率上，但该方法存在两方面局限性：第一，SRCNN 模型训练收敛太慢；第二，SRCNN 模型只适用于复原固定尺寸的图像[17]。为解决上述缺陷，Kim 等人[17] 通过使用深度网络结构中的多次级联小滤波器提出了 VDSR 模型，提升图像超分辨的性能。受 VGG[32]的启发，VDSR 也采用3×3 的级联小滤波器来高效地提取图像特征，提高图像的分辨率[1]。为了缓解梯度爆炸和梯度消失问题，VDSR 使用了梯度剪裁和残差学习操作。20 层的 VDSR 模型恢复高分辨率图像的流程如下：首先，VDSR 通过双三次插值操作放大输入图像，使其与参考图像相同尺寸；其次，使用 19 个堆积的卷积层来提取图像特征；最后，通过一层卷积来重建图像，获得高质量图像。其中，所有卷积核为3×3，第一层的输入通道为 3，最后一层的输出通道为 3，其余层的输入和输出通道都为 64。VDSR 的网络结构如图 3 所示。

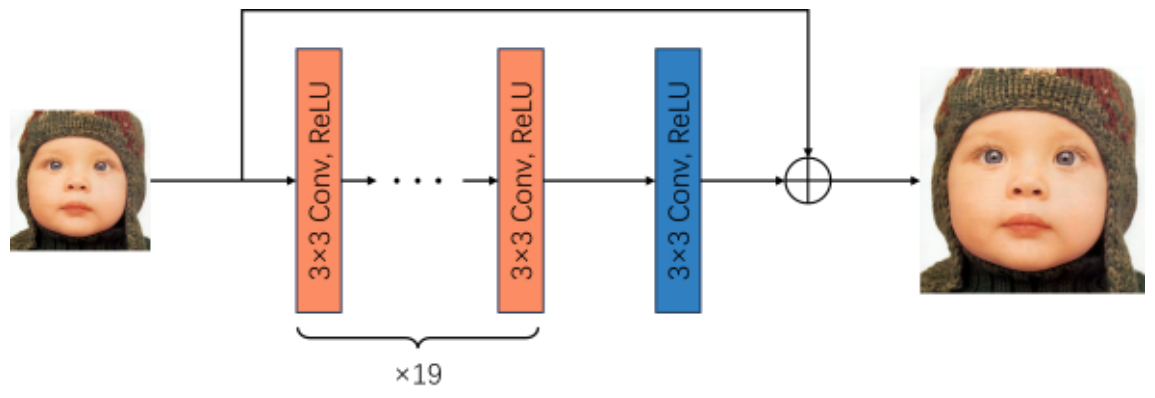

图 3 VDSR 的网络结构

**Fig.3 The Network Structure of VDSR**

虽然通过增加网络层数能提高图像超分辨的性能，但深度网络中浅层对深层作用会减少，这会导致模型收敛需要更多的训练数据[20]。为了克服这个挑战，Kim 等[20]采用递归操作改进 VDSR 实现 DRCN 图像超分辨网络，这不仅能降低梯度爆炸和梯度消失的影响，同时还能提升图像超分辨网络的表达能力。此外，将跳跃连接融合到网络中，加强浅层对深层的作用，增强网络的学习能力。例如，DRCN 采用感受野为41×41 的单卷积层作为递归单元，学习更精准的图像特征；并通过将每个递归单元的中间结果输入到重构模块，生成 HR 特征图来重建高分辨率图像。其中，递归单元后的一层卷积核为9×9，其余所有卷积核均为3×3，输入与输出的通道数均与 VDSR 一致。DRCN 的网络结构如图 4 所示。

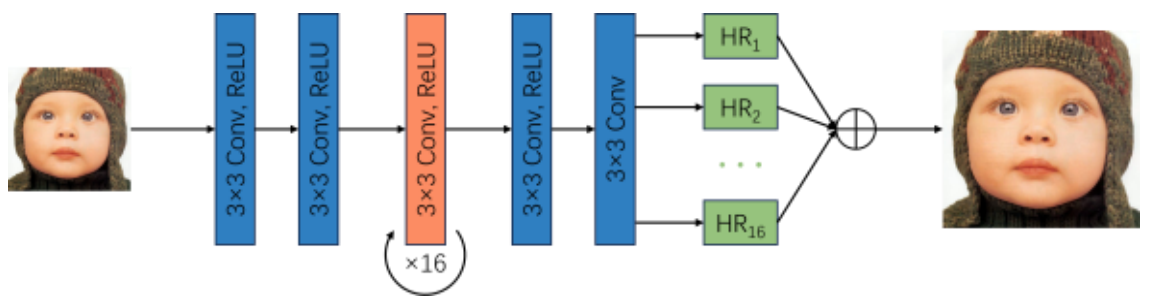

图 4 DRCN 的网络结构

**Fig.4 The Network Structure of DRCN**

DRCN 虽然将多重监督与递归单元结合在一起，在不引入大量参数的情况下，学习到更高层次的图像特征，但模型内部依然存在如下缺陷[18]：第一，采用双三次插值对图像预处理时，模型计算复杂度随高分辨率图像的空间大小呈二次增长。第二，模型非线性映射时，输入图像块依次被投影到

高维的 LR 与 HR 特征空间中，该过程计算昂贵且效率低下。为解决上述缺陷，Dong 等[18]提出了一种更简洁高效的网络结构 FSRCNN。该网络主要从三个方面改进超分辨模型：首先，在网络末端引入反卷积层，避免双三次插值的使用，这能直接从原始 LR 图像学习到 HR 图像的映射；其次，通过在映射前缩小输入特征维度，映射后再展开特征的方法来改进映射层，降低计算成本；第三，采用更小的过滤器尺寸，提高图像超分辨效率[1]，FSRCNN 的网络结构如图 5 所示。

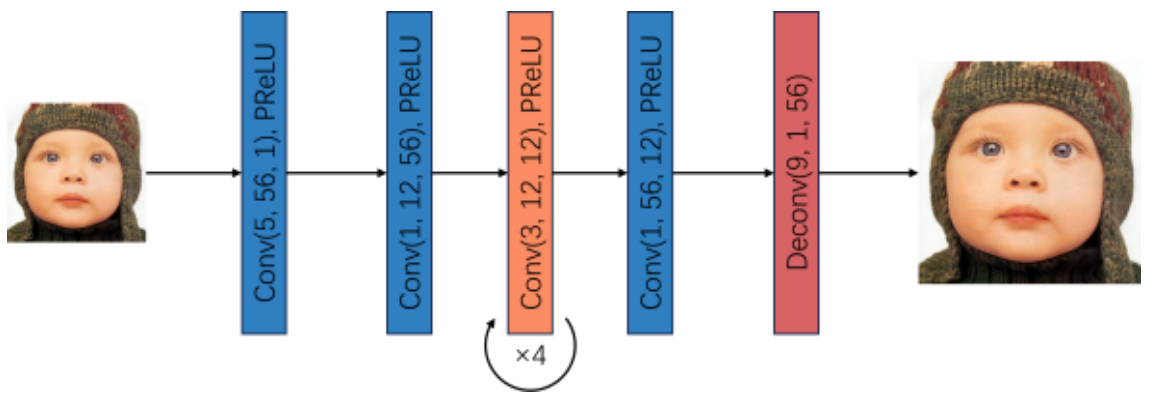

图 5　FSRCNN 的网络结构

**Fig.5 The Network Structure of FSRCNN**

尽管 FSRCNN 能降低模型的计算成本并提高了图像超分辨的质量，但在模型训练时，梯度消失与梯度爆炸的问题依然存在。随着卷积神经网络的设计趋于复杂化，模型的训练效率不断降低[21]。针对这一问题，Tong 等[21]提出了一种新颖的图像超分辨率方法，使用密集跳跃连接的图像超分辨网络（image super-resolution using dense skip connections, SRDenseNet）。为了缓解梯度消失与梯度爆炸的问题，SRDenseNet 首先采用密集跳跃连接操作嵌入到卷积神经网络中，有效结合低级特征与高级特征，改善网络中的信息流动[1]。为了降低训练复杂度，SRDenseNet 在特征提取时，允许模型重用前层特征映射信息，避免重复学习冗余特征，完成高分辨图像的重建。SRDenseNet 模型恢复 HR 图像的流程如下：输入图像依次通过卷积层学习低级特征、密集网络块学习高级特征、反卷积层学习上采样滤波器、重构层生成高分辨率的输出图像 [22]。SRDenseNet 的网络结构如图 6 所示。

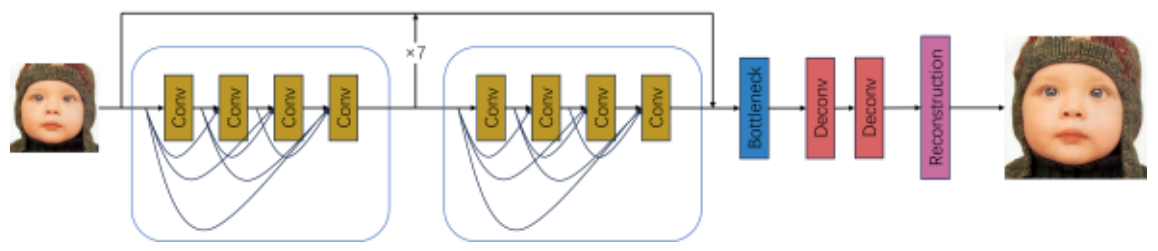

图 6　SRDenseNet 的网络结构

**Fig.6 The Network Structure of SRDenseNet**

随着网络深度的不断增加，SRDenseNet 引入的密集块虽然利用了低级和高级特征信息提高模型训练效率，但原始低分辨率图像的分层特征利用率不足，图像重建质量有待提高。为解决这一问题，Zhang 等[22]提出了 RDN。RDN 中的残差密集块（residual dense block, RDB）利用局部密集连接，结合网络所有层的特征信息，有效地提高了图像特征的利用率与图像超分辨效果。RDN 模型的图像超分辨流程如下：首先，输入图像通过卷积层来获得浅层的特征；随后，输入 3 个连续的 RDB 来提取深层特征；其次，利用密集操作融合网络中，结合网络内部所有层获取的特征信息，通过局部特征融合自适应增强特征；最后，使用反卷积和上采样操作得到高分辨率图像。其中，RDN 和 RDB 的网络结构如图 7 和 8 所示。

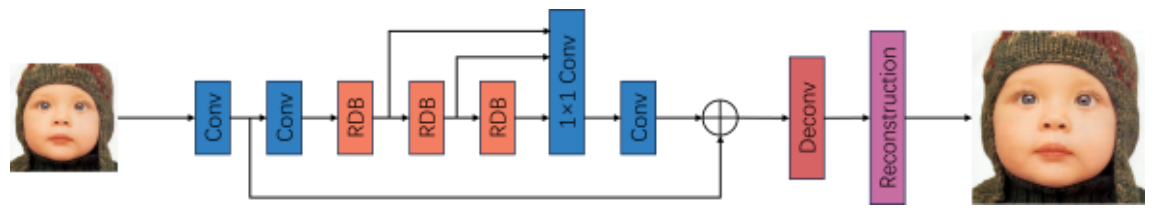

图 7　RDN 的网络结构

**Fig.7 The Network Structure of RDN**

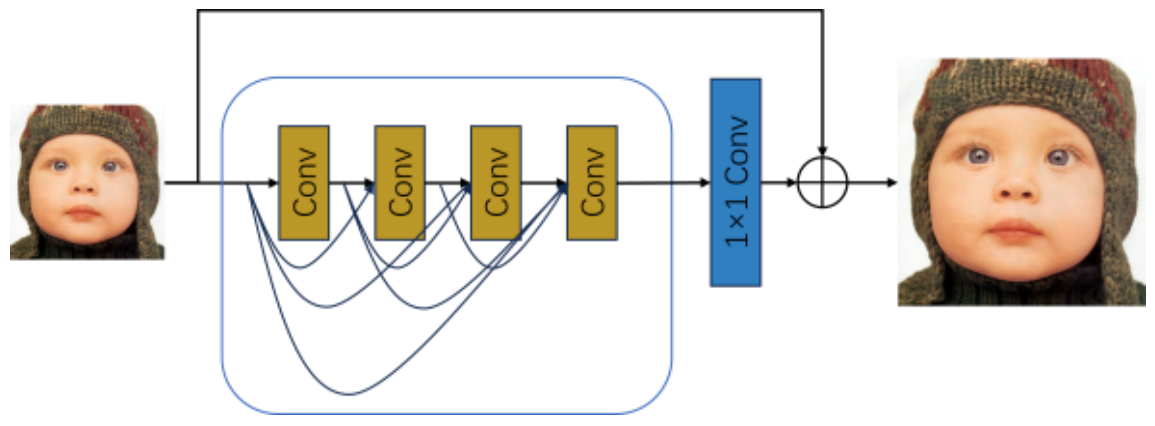

图 8　RDB 的网络结构

**Fig.8 The Network Structure of RDB**

## 2　面向图像超分辨的卷积神经网络方法

第 1 章按照技术演进的历程，已详细地介绍了一些面向图像超分辨的卷积神经网络的主流框架。为了进一步了解卷积神经网络在图像超分辨领域的发展与应用，本章将已有方法划分为基于插值方法和基于模块化方法。分类依据如下：插值算法是处理图像超分辨方法的经典技术，它能把低频信息转化到高频信息，是恢复高质量图像必不可少一步。随着端到端的卷积神经网络出现，插值算法已被用在卷积神经网络前（在网络输入前利用插值算法放大低分辨率图像）或网络末端（在网络深层利用插值算法放大低分辨率图像）用于放大获得低频信息，获得高清图像。然而，不同插值方法会影响图像超分辨方法的复原高质量图像性能，因此，本文根据将插值算法作为一种分类标准，并分为“双三次插值算法”、“最近邻插值算法”、“双线性插值算法”。由于标题字数关系，本文采用缩写，如“基

于双三次插值的卷积神经网络算法”写成“双三次插值算法”、“基于最近邻插值的卷积神经网络算法”写成“最近邻插值算法”、“基于双线性插值的卷积神经网络算法”写成“双线性插值算法”。插值方法放大图像时容易丢失细节信息，由于卷积神经网络在训练过程中具有动态学习参数，因此，学者们将卷积和插值方法结合提出了模块化方法，提高图像超分辨鲁棒性。每种方法包括三种典型计算类型，按照图像退化类型是否已知，从图像超分辨与图像盲超分辨的角度，介绍了多种图像超分辨率经典算法、设计思路和实际应用。本章结构如图 1 所示。

## 2.1 基于插值的卷积神经网络图像超分辨方法

随着显示器等硬件设备的迅速发展，一张图像通常需要以不同的尺寸和不同的纵横比显示。插值操作是常见的缩放图像，可以根据待补充像素点的周围信息，确定缺失像素。所以将插值操作应用于神经网络的预上采样中，可以初步补充像素点，扩大待处理图像的分辨率和尺寸，使与参考的图像具有相同大小。此外，通过修改输入图像的设置，如大小、分辨率及横纵比等，使卷积神经网络更容易建立输入图像与输出图像的映射[33]。因此，卷积神经网络在处理图像超分辨任务时会采用不同的图像插值技术[34]，例如双三次插值[35]、最近邻插值[4]和双线性插值[36]等来缩放待处理的图像的尺寸，使得与参考的图像具有相同的尺寸，利于训练图像超分辨模型。本节将基于插值的卷积神经网络图像超分辨方法根据插值方法的不同分为三类：双三次插值算法、最近邻插值法和双线性插值法，并分别介绍它们在图像非盲和盲超分辨率任务上的应用，本小节结构如图 9 所示。

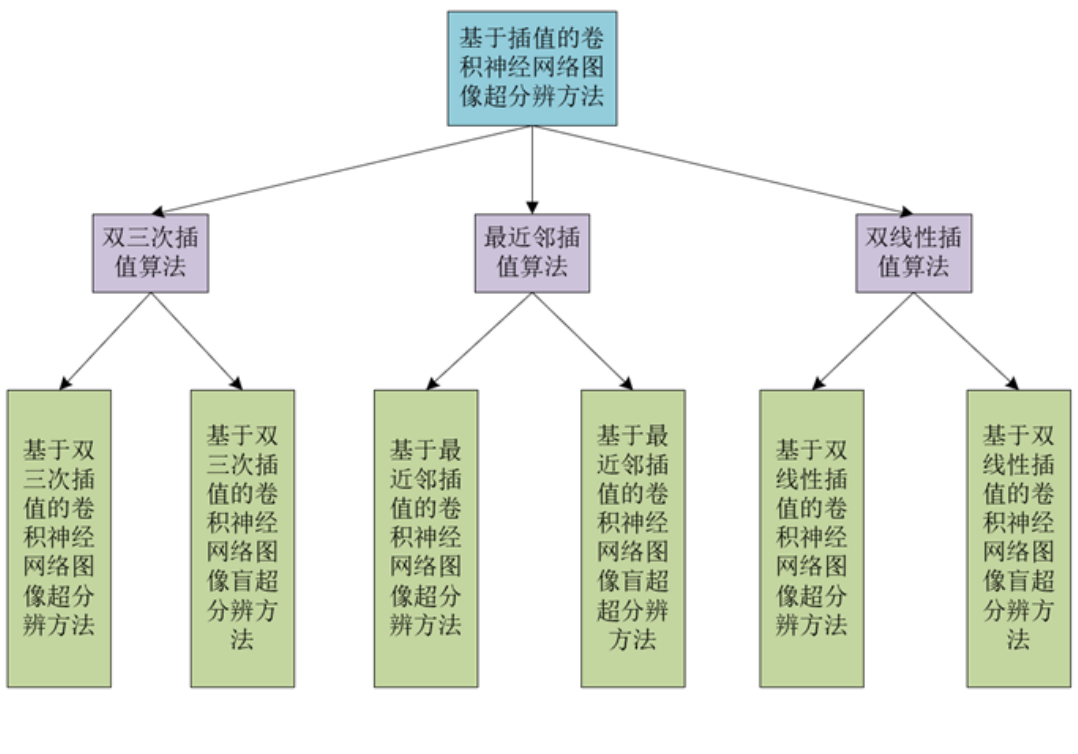

图 9　基于插值的卷积神经网络图像超分辨方法分类结构图

**Fig.9 The Classification Structure of Interpolation-Based Convolutional Neural Network Methods for Image Super-Resolution**

### 2.1.1 双三次插值算法

1）基于双三次插值的卷积神经网络图像超分辨方法

双三次插值方法主要通过计算未知像素点周围 16 个像素的加权平均值，作为最终的插值结果。由于周围像素与未知像素的距离各不相同，像素距离越近，权重越高，该方法插值效果较好。

基于以上双三次插值方法原理，卷积神经网络将其用在图像超分辨上。具体为，首先利用双三次插值方法将低分辨率图像扩大成与参考的高质量图像相同尺寸大小的图像，其次，将上步获得的图像通过卷积神经网络来预测对应的高质量图像。例如：Dong 等[16]通过双三次插值操作来预处理给出的低分辨率图像，并通过三层的卷积神经网络以像素点的方式映射得到高质量图像。虽然该方法在图像超分辨上比传统的机器学习方法获得更有效的效果，但该网络可扩展性差，其性能随着网络深度增加而下降。为了克服这些缺点，VDSR 采用更小的卷积核以堆积的方式增大网络深度，扩大感受野信息，提取更多的上下文信息。并通过残差操作来增强浅层特征的作用，提升图像超分辨的性能[37]。为了充分利用网络的结构信息，Ren 等[38]通过增加网络宽度扩大感受野，实现了基于上下文网络融合的卷积神经网络图像超分辨方法（Context-wise Network Fusion, CNF），提高恢复图像的质量。其中，CNF 通过并行三个层数不同的 SRCNN 获得三幅不同的高质量图像，融合它们能获得重建图像[39]。为了获得互补的信息，Hu 等[40]提出级联多尺度交叉卷积神经网络（cascaded multi-scale cross，CMSC）。其中，CMSC 采用双三次插值来放大低分辨率图像，使其与参考图像具有相同的大小；随后，采用级联子网络以由粗到细的方式学习精准的高频特征；为了增强网络的表达能力，将残差学习操作融合到 CMSC 中；最后，采用多尺度交叉模块提取互补的多尺度信息，保证提取特征的鲁棒性，提高图像超分辨率的性能。为了降低计算代价，Yamanaka 等[41]提出了基于残差学习跳跃连接的深度卷积神经网络（deep CNN with residual net, skip connection and network, DCSCN）。DCSCN 将跳连接嵌入到深度网络中，提取局部和全局的图像特征；为了减少网络的复杂度，采用两个 $1\times1$ 卷积对获得特征进行降维，最后通过一个 $3\times3$ 的卷积进行重构和获得预测的 HR 图像。

尽管上述模型提高了模型训练效率，但仍然存在如下问题：LR 图像的层次特征利用不足、模型过大等。为提高图像特征利用率，RDN 利用 RDB 实现图像局部特征抽取、全局特征融合，充分利用了 LR 图像的多层次特征。而为了解决上述模型由于层数过多导致模型过大的问题，例如，过拟合、存储与检索困难等[42]。DRCN 率先将递归神经网络引入图像超分辨领域，通过卷积层间权重共享，在不增加额外参数的条件下，恢复了更多的图像信息，降低了梯度爆炸和梯度消失的影响。为了降低模型参数，提高训练效率，DRRN[43]在 DRCN 的基础上改进，不同于卷积层间共享权重，DRRN 将多个残差单元组成递归块，在递归块间共享权重，显著提高图像复原精度，实现了深度而简洁的网络。同样利用残差学习提高图像质量的超分辨模型还包括增强的深度卷积神经网络（enhanced deep super-resolution network, EDSR）、多尺度深度卷积神经网络（multi-scale deep super-resolution system, MDSR）[44]、基于残差通道注意力的卷积神经网络（residual channel attention network, RCAN）[23]等。

为了解决深度网络容易出现梯度消失或梯度下降的难题，Tai 等[39]提出基于记忆网络的卷积神经网络（memory network, MemNet）。MemNet 利用连接操作将浅层信息传递到深层，这不仅能缓解深度网络中梯度消失或梯度爆炸问题，还能提高图像超分辨率效果[45]。Tian 等[46]提出由粗到细卷积神经网络的图像超分辨方法（coarse-to-fine super-resolution CNN, CFSRCNN）。该方法分别提取低频和高频信息，并将连接和残差学习作用到网络的层次信息上，融合高频和低频信息，解决由深层上采样操作突然放大获得的特征，导致网络训练难的问题。

为了解决真实世界的图像增强问题，Rad 等[47]使用 GAN 实现了一个真实双三次超分辨率的卷积神经网络（real bicubic super-resolution, RBSR）。具体为，首先使用一个基于卷积神经网络的生成器将真实的低分辨率图像转换为类似于双三次下采样的低分辨率图像。随后，通过预训练的 EDSR[44]网络作用在获得低分辨率图像上，预测高质量图像。此外，基于数字高程模型的超分辨率生成对抗网络（digital elevation model-SRGAN, D-SRGAN）[48]和基于高分辨率表征学习的医学图像超分辨率生成对抗网络（GAN-based medical image SR network via HR representation learning, Med-SRNet）[49]能有效地获得图像的细节信息，提高真实图像的分辨率。为了方便读者理解不同超分辨方法的原理和区别，表 1 通过作者、方法、提出时间、应用和关键字来展示基于双三次插值卷积神经网络方法的图像超分辨方法。

**表 1　基于双三次插值的图像超分辨卷积神经网络方法**

**Table 1 Image Super-Resolution Convolutional Neural Network Method Based on Bicubic Interpolation**

| 作者 | 方法 | 提出时间 | 应用 | 关键字 |
|---|---|---|---|---|
| Dong 等 | SRCNN[16] | 2015 年 | 图像超分辨 | 基于像素映射的卷积神经网络 |
| Kim 等 | VDSR[17] | 2016 年 | 图像超分辨 | 深度的卷积神经网络 |
| Ren 等 | CNF[38] | 2017 年 | 图像超分辨 | 基于上下文网络融合的卷积神经网络 |
| Hu 等 | CMSC[40] | 2018 年 | 图像超分辨 | 级联多尺度交叉的卷积神经网络 |
| Kim 等 | DRCN[20] | 2016 年 | 图像超分辨 | 基于深度递归的卷积神经网络 |
| Tai 等 | DRRN[43] | 2017 年 | 图像超分辨 | 基于深度递归的残差卷积神经网络 |
| Tai 等 | MemNet[39] | 2017 年 | 图像超分辨 | 基于记忆网络的卷积神经网络 |
| Lim 等 | EDSR[44] | 2017 年 | 图像超分辨 | 增强的深度卷积神经网络 |
| Lim 等 | MDSR[44] | 2017 年 | 图像超分辨 | 多尺度深度卷积神经网络 |
| Wang 等 | ProSR[50] | 2018 年 | 图像超分辨 | 基于逐层上采样的卷积神经网络 |
| Zhang 等 | RCAN[23] | 2018 年 | 图像超分辨 | 基于残差通道注意力的卷积神经网络 |
| Ni 等 | CGDMSR[51] | 2017 年 | 图像超分辨 | 基于深色和彩色图像的卷积神经网络 |
| Jin 等 | DCSCN[41] | 2017 年 | 图像超分辨 | 基于残差学习的深度卷积神经网络 |
| Han 等 | SSF-CNN[52] | 2018 年 | 图像超分辨 | 基于空间和光谱的卷积神经网络 |
| Ledig 等 | SRGAN[25] | 2017 年 | 图像超分辨 | 生成对抗的网络 |
| Wang 等 | LISTA[53] | 2015 年 | 图像超分辨 | 基于稀疏编码的卷积神经网络 |
| Hui 等 | IDN[54] | 2018 年 | 图像超分辨 | 基于信息蒸馏的卷积神经网络 |
| Tian 等 | CFSRCNN[46] | 2020 年 | 图像超分辨 | 由粗到细卷积神经网络 |
| Rad 等 | RBSR[47] | 2021 年 | 图像超分辨 | 基于真实双三次超分辨率的卷积神经网络 |
| Demiray 等 | D-SRGAN[48] | 2021 年 | 图像超分辨 | 基于数字高程模型的超分辨生成对抗网络 |
| Zhang 等 | Med-SRNet[49] | 2022 年 | 图像超分辨 | 基于高分辨率表征学习的医学图像超分辨率生成对抗网络 |

2）基于双三次插值的卷积神经网络图像盲超分辨方法

大多数基于 CNN 方法都通过已知先验来建立图像退化模型。然而，当真实捕获到图像的退化类型与预先假设不同时，会降低图像复原方法的性能。例如，为解决盲图像超分辨率中未知模糊核的估计问题，Yuan 等[55]提出了基于循环-循环的生成对抗网络（cycle-in-cycle GAN, CinCGAN）。其中，CinCGAN 包含两个 CycleGAN，第一个 CycleGAN 将各种降质因素的低分辨图像映射到干净的双三次下采样的低分辨率空间中，实现去噪和去模糊。通过预训练的深度模型将中间结果上采样到所需的高分辨大小图像。最后，通过第二个 GAN 对第一个 CycleGAN 进行微调和优化，生成高质量高分辨的图像。为了更好处理不同模糊核的低分辨图像，Maeda 等[56]通过改进 CycleGAN 实现非成对的图像超分辨率卷积神经网络（unpaired image super-resolution, UISR）。UISR 仅使用非配对的核/噪声校正网络来去除噪声和模糊核，生成干净的LR 图像。为了重建高分辨率图像，该方法利用伪配对的超分辨率网络学习到从伪干净 LR 图像到原始高分辨率（HR）图像的映射，实现超分辨率图像重建功能。Zhang 等[57]提出了深度即插即用的超分辨卷积神经网络（deep plug-and-play super-resolution, DPSR）。该方法通过设计一个新的单图像超分辨率退化模型和利用半二次分裂算法优化能量函数，实现处理任意模糊核的低分辨率图像功能。为了提高模糊核的估计精度，Gu 等[58]提出了基于迭代内核校正的卷积神经网络（iterative kernel correction, IKC）。IKC 能根据已有的 SR 结果对估计核进行迭代修正，逐渐恢复逼近真实 HR 图像来去除因内核不匹配导致的图像伪影。因 IKC 只关注有限类型的模糊核与噪声。因此，该方法并不是完全意义上的图像盲超分辨率方法。

为了提高真实图像盲超分辨率方法的泛化性和鲁棒性，Zhou 等[59]提出了内核建模超分辨率卷积神经网络（kernel modeling super-resolution, KMSR）。KMSR 主要分为两个阶段：首先，使用生成对抗网络构建了一组真实的模糊核，随后使用生成的核构建了 HR 和相应的 LR 图像。之后，利用它们来训练超分辨率网络，提升了真实照片盲超分辨率方法的性能。为了处理复杂场景下真实图像的盲超分辨率，Kligler 等[60]提出了内核生成对抗网络方法，该方法仅利用 LR 图像进行训练代替已知的先验知识，学习其内部的图像块分布来估计出模糊核，实现自适应的下采样操作，提高了盲超分辨模型的鲁棒性。为了真实世界中未知退化的图像超分辨问题，Wang 等人[61]提出基于退化感知的卷积神经网络超分辨方法（degradation-aware super-resolution, DASR）。DASR 从退化表示中预测卷积核和信道级调制系数，增强图像超分辨模型的鲁棒性。为了更有效地处理不同退化情况下的图像超分辨问题，Zhang 等人[62]引入随机洗牌策略随机学习模糊、下采样和噪声实现图像盲复原模型。为了提高真实图像的盲超分辨效率，Wang 等人[45]利用高阶退化引导 GAN 网络，实现图像盲超分辨模型。表 2 概述了更多基于双三次插值的单图像盲超分辨率卷积神经网络。

**表 2　基于双三次插值卷积神经网络的图像盲超分辨方法**

**Table 2 Image Blind Super-Resolution Method Based on Bicubic Interpolation Convolutional Neural Network**

| 作者 | 方法 | 提出时间 | 应用 | 关键词 |
|---|---|---|---|---|
| Yuan 等 | CinCGAN[55] | 2018 年 | 图像盲超分辨 | 基于循环中循环的生成对抗网络 |
| Maeda 等 | UISR[56] | 2020 年 | 图像盲超分辨 | 非成对图像生成对抗网络 |
| Zhang 等 | DPSR[57] | 2019 年 | 图像盲超分辨 | 深度即插即用超分辨卷积神经网络 |
| Gu 等 | IKC[58] | 2019 年 | 图像盲超分辨 | 基于迭代内核校正的卷积神经网络 |
| Zhou 等 | KMSR[59] | 2019 年 | 图像盲超分辨 | 内核建模卷积神经网络 |
| Kligler 等 | KernelGAN[60] | 2019 年 | 图像盲超分辨 | 内核生成对抗网络 |
| Wang 等 | DASR[61] | 2021 年 | 图像盲超分辨 | 基于退化感知的超分辨率卷积神经网络 |
| Zhang 等 | Blind ESRGAN[62] | 2021 年 | 图像盲超分辨 | 增强的盲超分辨卷积神经网络 |
| Wang 等 | Real ESRGAN[45] | 2021 年 | 图像盲超分辨 | 增强的真实图像超分辨卷积神经网络 |
| Zhang 等 | CRL-SR[63] | 2021 年 | 图像盲超分辨 | 基于对比表示学习的盲图像超分辨率卷积神经网络 |
| Wu 等 | CDCN[64] | 2022 年 | 图像盲超分辨 | 基于组件分解与协同优化的卷积神经网络 |
| Cao 等 | PCNet[65] | 2021 年 | 图像盲超分辨 | 基于先验校正网络的盲超分辨率卷积神经网络 |
| Yamawaki 等 | BSR-DUL[66] | 2021 年 | 图像盲超分辨 | 基于联合优化策略的盲超分辨率卷积神经网络 |
| Yamawaki 等 | DIP[67] | 2021 年 | 图像盲超分辨 | 基于可学习退化模型的盲无监督学习卷积神经网络 |

### 2.1.2 最近邻插值算法

1）基于最近邻插值的卷积神经网络图像超分

辨方法

最近邻插值算法通过直接选择待插入点最近的相邻像素作为插入点的像素值来完成插值操作，完成图像超分辨任务[4]。虽然该方法能在一定程度上提高计算速度，但该过程仅复制已有的像素值，这会导致在放大图像时失真较为严重。该算法与其他插值算法相比，思想最简单，计算速度最快，也更容易实现，因此它已广泛应用于卷积神经网络中解决图像超分辨问题[68]。

早期的卷积神经网络通过反卷积实现把低频特征转为高频特征，获得高质量图像。其中，反卷积层是基于最近邻近插值算法实现的[18]，例如，FSRCNN 利用深层的反卷积层从获得 LR 特征图转化为 SR 特征图，重构高质量图像[18]为了提高图像超分辨模型的适应能力，Zhang 等[69]提出超分辨卷积神经网络（unfolding super-resolution network, USRNet）。USRNet 通过半二次分裂算法展开 MAP 推理，得到固定次数的迭代计算，用于交替求解一个数据子问题和一个先验子问题；随后，这两个子问题由神经模块解决，得到一个端到端的迭代网络。为了降低计算成本和模型复杂度， Fan 等[70]通过深度约束的残差设计提出基于平衡两阶段残差的卷积神经网络（balanced two-stage residual networks, BT-SRN），在图像超分辨任务的性能与效率均取得好的效果。为了防止峰值信噪比不能全面地衡量图像的质量问题，Sajjadi 等[71]提出了增强网络（EnhanceNet），该网络采用对抗训练，并通过结合像素级损失，感知损失和纹理匹配损失生成高质量、具有逼真纹理的超分辨率图像。

为了降低超分辨图像中的伪影以及失真细节，增强型超分辨率生成对抗网络（enhanced super-resolution generative adversarial networks, ESRGAN）[26]引入残差密集块和相对平均鉴别器，并改进感知损失和对抗性训练来改进 SRGAN 网络，解决图像超分辨问题。其中，残差密集块增加了网络中层次之间的关系，增强深度网络的结构信息；其次，在权重初始化过程中采用了更小的初始化值来提高深度网络训练的稳定性；最后，考虑到当前生成图像的真实度，使用相对均值鉴别器来增加生成图像的更多真实细节信息，这能减少生成图像与平均真实图像之间的相对差异，使其更加符合人类视觉感知。由于不同图像的纹理细节差异较大，图像超分辨率方法很难提取不同图像中的细节信息。为了解决这个问题，Shang 等[72]提出了基于感受野块的增强超分辨率生成对抗网络（super resolution network with receptive field block based on Enhanced SRGAN, RFB-ESRGAN）。RFB-ESRGAN 利用小卷积核代替大卷积核提取不同尺度的特征信息来提取更多细节信息，获得更多图像细节，提高图像超分辨性能最后，该网络交替使用侧重于空间维度的最近邻插值和侧重于深层维度的亚像素卷积执行上采样，实现深层信息和空间信息充分交流，达到生成更多图像细节的效果。为了降低模型复杂度， Zhao 等[73]提出了像素注意力卷积神经网络（pixel attention network, PAN）。PAN 利用像素级注意力模块对输入特征图以加权的方式来突出关键像素，抑制无关信息，并结合残差块和多尺度特征融合，优化网络结构，在减少计算复杂度同时提升图像超分辨性能。表 3 展示了更多基于最近邻插值的单图像超分辨卷积神经网络的方法。

**表 3　基于最近邻插值的单图像超分辨率卷积神经网络**

**Table 3 Image Super-Resolution Method Based on Nearest Neighbor Interpolation Convolutional Neural Network**

| 作者 | 方法 | 提出时间 | 应用 | 关键词 |
|---|---|---|---|---|
| Dong 等 | FSRCNN[18] | 2016 年 | 图像超分辨 | 快速超分辨卷积神经网络 |
| Zhang 等 | USRNet[69] | 2020 年 | 图像超分辨 | 展开超分辨率卷积神经网络 |
| Fan 等 | BT-SRN[70] | 2017 年 | 图像超分辨 | 基于平衡两阶段残差的卷积神经网络 |
| Sajjadi 等 | EnhanceNet [71] | 2017 年 | 图像超分辨 | 基于感知损失、对抗损失和纹理损失的增强卷积神经网络 |
| Wang 等 | ESRGAN[26] | 2019 年 | 图像超分辨 | 增强的超分辨率生成对抗网络 |
| Shang 等 | RFB-ESRGAN[72] | 2020 年 | 图像超分辨 | 基于感受野块的增强超分辨率生成对抗网络 |
| Zhao 等 | PAN[73] | 2020 年 | 图像超分辨 | 像素注意力卷积神经网络 |
| Huang 等 | CCNet[74] | 2019 年 | 图像超分辨 | 基于跨尺度通信的卷积神经网络 |
| Jo 等 | LUT[75] | 2021 年 | 图像超分辨 | 基于查找表的实用图像超分辨率卷积神经网络 |

2）基于最近邻插值的卷积神经网络图像盲超分辨方法

现有图像盲超分辨方法均基于模糊核具有空间不变性的假设,但由于物体运动和失焦等因素,模糊核通常是空间变化的,这些图像盲超分辨模型在实际应用中性能较差。为解决这个问题，Liang 等[76]

提出相互仿射的卷积神经网络（mutual affine network, MANet）。MANet 引入一种全新的互仿射卷积层,使得网络能在不增加感受野、模型大小和计算负担的前提下增强特征表达能力,使得图像超分辨模型在模糊核空间变化中展现了优秀的性能。表 4 展示了更多的基于最近邻插值卷积神经网络的图像盲超分辨方法。

**表 4　基于最近邻插值卷积神经网络的图像盲超分辨方法**

**Table 4 Image Blind Super-Resolution Method Based on Nearest Neighbor Interpolation Convolutional Neural Network**

| 作者 | 方法 | 提出时间 | 应用 | 关键词 |
|---|---|---|---|---|
| Zhang 等 | Blind ESRGAN[62] | 2021 年 | 图像盲超分辨 | 增强的盲超分辨卷积神经网络 |
| Liang 等 | MANet[76] | 2021 年 | 图像盲超分辨 | 相互仿射的卷积神经网络 |
| He 等 | SRDRL[77] | 2021 年 | 图像盲超分辨 | 基于退化重构损失的超分辨卷积神经网络 |

#### 2.1.3 双线性插值算法

1）基于双线性插值的卷积神经网络图像超分辨方法

双线性插值算法加权平均待插入点四个邻域像素，计算最终插入值。具体而言，它在水平和垂直两个方向上分别计算插值权重，再结合四个邻域像素的值，从而得到插入点的估计值。最近邻方法没有考虑不同方向的信息，而双线性算法保留水平和垂直的边缘信息，因此使得图像变得更平滑，应用在图像超分辨任务中。例如，Youm 等[78]将多通道输入引入到卷积神经网络中提出了多通道输入的卷积神经网络（multi-channel-input super-resolution convolutional neural networks, MC-SRCNN），提高了网络的泛化能力，提高图像超分辨性能。具体为：MC-SRCNN 利用双线性插值、双三次插值通道等作为多通道输入，这不仅能有效减轻 SRCNN 中梯度消失和梯度保障来提高训练稳定性，还使得网络获得更丰富的信息，提高 SRCNN 性能。为了解决 SRCNN 恢复高分辨率的图像效果不佳问题时，Intaniyom 等[79] 通过双阶段处理模块来提高在图像超分辨任务放大过程中预测高质量，而提出的改进超分辨率卷积神经网络（modified super-resolution CNN, m-SRCNN）。其中，第一阶段通过对图像进行锐化、去噪、缩放和裁剪处理，粗略地获得相对干净的低分辨率图像。第二阶段利用图像增强模块、放大和增强模块、双重增强模块和双重放大模块从不同方面来增强图像，提高预测图像的分辨率。为了降低计算成本和减少模型复杂度，学者们提出了很多优化图像超分辨网络的方法。例如，Vu 等[80]通过使用通道增加网络深度，称为快速、高效的增强卷积神经网络（fast and efficient quality enhancement, FEQE），这不仅可以减少计算复杂度，还可以提取更多特征信息，提高图像保真度。为了提高图像超分辨模型的训练效率，Liu 等[81]设计渐进残差块来对深度特征逐步下采样来提出渐进式残差学习的卷积神经网络（progressive residual learning for single image SR, PRLSR），减少冗余信息，获得更精准的结构信息，这能减少图像的细节损失。此外，可学习权值的残差结构，不仅可以提取更多层的特征，还能自适应调整残差映射与恒等映射在残差结构中影响，加快收敛速度，提高图像超分辨性能。为了克服卷积操作引起像素偏移问题，Zhang 等[82]将可变形卷积引入到深度残差网络中，提出了可变形的残差卷积神经网络（deformable and residual convolutional network, DefRCN），解决图像超分辨问题。DefRCN 采用自适应调整采样网络，学习更精准的空间 信息，解决了固定卷积核大小限制卷积神经网络在图像超分辨性能的问题。此外，将改进的残差卷积块融合到卷积神经网络中，不仅能提高训练效率，也能防止深度网络的梯度消失问题。为了解决未知退化问题，Kim 等[83]基于 CinCCGan[55]提出了循环的生成对抗网络（cycle GAN）。该网络结合像素级损失、感知损失、结构相似损失、区域鉴别器来根据目标区域生成图像，解决未有参考高分辨率图像的图像超分辨问题。表 5 介绍了更多基于双线性插值算法的图像超分辨率卷积神经网络。

**表 5　基于双线性插值的单图像超分辨率卷积神经网络**

**Table 5 Image Super-Resolution Convolutional Neural Network Based on Bilinear Interpolation**

| 作者 | 方法 | 提出时间 | 应用 | 关键词 |
|---|---|---|---|---|
| Youm 等 | MC-SRCNN[78] | 2016 年 | 图像超分辨 | 多通道输入的卷积神经网络 |
| Intaniyom 等 | m-SRCNN[79] | 2021 年 | 图像超分辨 | 改进的超分辨率卷积神经网络 |
| Vu 等 | FEQE[80] | 2019 年 | 图像超分辨 | 快速且高效的增强卷积神经网络 |
| Liu 等 | PRLSR[81] | 2020 年 | 图像超分辨 | 渐进式残差学习的卷积神经网络 |
| Kim 等 | cycle GAN[83] | 2020 年 | 图像超分辨 | 循环中循环的生成对抗网络 |
| Fan 等 | SCN[84] | 2020 年 | 图像超分辨 | 多尺度卷积神经网络 |
| Zhang 等 | DefRCN[82] | 2022 年 | 图像超分辨 | 可变形与残差卷积神经网络 |
| Suryanarayana 等 | VDR-net[85] | 2021 年 | 图像超分辨 | 深度残差卷积神经网络 |
| Umer 等 | SR2GAN[86] | 2021 年 | 图像超分辨 | 深度超分辨率残差生成对抗网络 |

2）基于双线性插值的卷积神经网络图像盲超分辨方法

为了提高核估计效果， Zhu 等[87]提出了基于编码器-解码器的内核估计卷积神经网络（encoder-decoder based kernel estimation, EDKE）。EDKE 首先通过编码器网络学习真实图像的退化过程，随后，利用解码器将 LR 图像放大到输入大小，并映射到原始图像本身，解决了核各向异性限制了核估计在实践中的精确值问题，实现了更准确的核估计与图像复原。为了提高真实图像的盲超分辨复原效果，Wei 等[88]提出领域距离感知的超分辨率卷积神经网络（domain-distance aware super-resolution, DASR）。该网络采用领域距离感知训练和领域距离加权监督策略，能有效地减少训练数据和测试数据之间的域间差距，使得输出图像更自然和真实，提高本方法在真实世界中的实用性。表 6 概述了上述基于双线性插值的卷积神经网络的图像盲超分辨方法。

**表 6　基于双线性插值的单图像盲超分辨率卷积神经网络**

**Table 6 Image Blind Super-Resolution Methods Based on Bilinear Interpolation Convolutional Neural Network**

| 作者 | 方法 | 提出时间 | 应用 | 关键词 |
|---|---|---|---|---|
| Zhu 等 | EDKE[87] | 2021 年 | 图像盲超分辨 | 基于编码器-解码器的内核估计卷积神经网络 |
| Wei 等 | DASR[88] | 2021 年 | 图像盲超分辨 | 领域距离感知的超分辨率卷积神经网络 |

## 2.2 基于模块化的卷积神经网络图像超分辨方法

由于卷积神经网络具有端到端结构,用软件工程中模块化设计的思想引导卷积神经网络成为解决图像处理任务的主流方法[89]。而在图像超分辨中,用于提高分辨率的转置卷积,亚像素层和元上采样操作也被封装成模块来完成图像超分辨任务。根据不同上采样模块可以将基于模块化的卷积神经网络图像超分辨方法分为基于转置卷积模块的图像超分辨,基于亚像素层模块的图像超分辨,基于元上采样模块的图像超分辨方法,具体如图 10 所示：

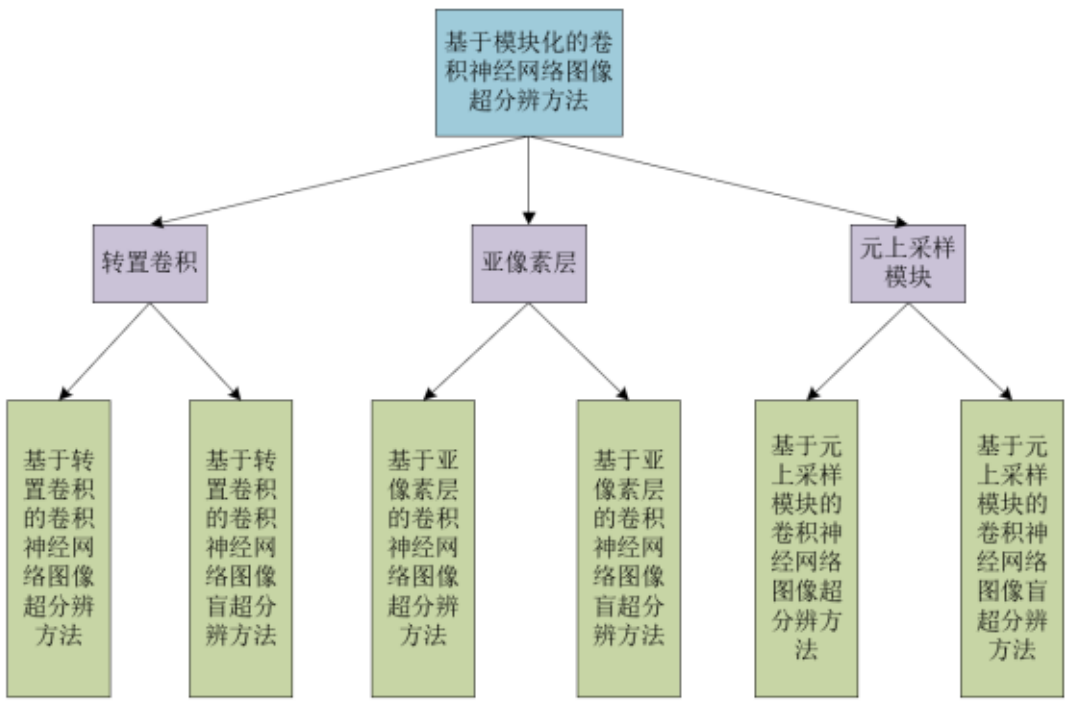

**图 10 基于模块化的卷积神经网络图像超分辨方法分类结构图**

**Fig.10 The Classification Structure Diagram of Modular Convolutional Neural Network Methods for Image Super-Resolution**

### 2.2.1 转置卷积（反卷积）

1）基于转置卷积模块的卷积神经网络图像超分辨方法

转置卷积与插值的上采样方法在图像超分辨中的作用相同，都是用来把低分辨率图像放大成高分辨率图像。但他们的原理不同，转置卷积先对低分辨率图像进行零填充，也就是在像素之间填充零像素，得到扩大的特征图，后在过大的特征图上进行卷积操作以重构出高分辨率图像。由于转置卷积在卷积核中自动学习权重，不需要预定义的插值权重或插值操作，它比插值方法的效率更高。例如，Li 等[89]整合特征映射的反卷积、多尺度融合和残差学习，实现高效的超分辨网络。为了提高模型训练效率，Mao 等[90]提出深度残差的编码器-解码器卷积神经网络（Very Deep Residual Encoder-Decoder Networks, RED-Net）。RED-Net 由多层卷积和反卷积算子组成。首先，该网络使用跳跃连接来对称连接卷积层和反卷积层，跳跃连接允许信号反向传播到底层，解决梯度消失的问题，使得训练深度网络更加容易。随后，再次利用跳跃连接将图像细节从卷积层传递到反卷积层，提高图像恢复性能。该网络使用跳跃连接来对称链接卷积层和反卷积层，跳跃连接允许信号反向传播到底层，解决梯度消失的问题，使得训练深度网络更加易训练。随后，再次利用跳跃连接将图像细节从卷积层传递到反卷积层，提高图像恢复性能。Lai 等[91]提出了拉普拉斯金字塔超分辨率卷积神经网络（Laplacian pyramid super-resolution network, LapSRN）。LapSRN 由特征提取分支和图像重构分支组成，并将图像分解成不同分辨率的层级，每个层级包含不同频率的信息，其具体超分辨流程如下。首先，网络从 LR 图像中提取特征，随后，在每个金字塔层级学习当前分辨率图像与 HR 图像间的差异。最后，将学习的差异信息添加到上采样后的图像中生成高分辨率输出图像并作为下一层级的输入。通过这种逐级处理，逐层提高图像的分辨率和细节质量最终完成超分辨率重建。

为了使得深度网络更适用于真实的手机相机的图像复原问题，Hui 等[54]提出信息蒸馏的卷积神经网络（information distillation network, IDN）来解决图像超分辨问题。IDN 是由特征提取块、信息蒸馏块和重构块来实现轻量级的卷积神经网络，它的具体工作原理如下：首先，利用特征提取块初步地提取图像特征；随后，信息蒸馏块通过增强单元和压缩单元逐层提取和压缩特征信息，将长短路径特征融合以获得更多的有效信息。最后，重建块利用残差学习操作，融合不同层获得特征，提高获得特征的鲁棒性，提高图像超分辨性能。同样改进残差学习的超分辨网络还有 Shi 等[92]提出具有固定跳跃连接的广泛残差卷积神经网络（wide residual network with a fixed skip connection, FSCWRN），该网络结合全局残差学习和基于浅层网络的局部残差学习，采用渐进式宽网络代替深层网络，缓解了深度网络存在的退化和特征重用减少的问题。尽管跳跃连接缓解了梯度爆炸和梯度消失问题，但是跳跃连接仍然存在缺陷：模型前层无法从后层访问有效信息。针对这个问题，Li 等[93]提出了超分辨率反馈卷积神经网络（super-resolution feedback network, SRFBN）。SRFBN 利用带有约束的循环神经网络中的隐藏状态实现反馈机制，通过反馈连接自上而下地传输高级信息，低级信息通过高级信息的反馈得到逐步的细化和增强，最终生成高分辨率图像。

为了提高超分辨模型在大幅图像的运行速度和恢复质量。Zeng 等[94]提出了多对多连接的卷积神经网络（many-to-many connection of network，MMCN）。在重构过程中，MMCN 采用反卷积层和局部多态并行网络，提高图像恢复质量。为了处理较大比例因子缩放的图像超分辨问题，Haris 等人[95]提出深度反向投影卷积神经网络（Deep back-projection network，DBPN）。该网络创新地结合迭代的上采样和下采样层，通过将 HR 特征投影回 LR 空间的误差反馈机制增强特征表示和重建精度，提高图像超分辨效果。为了在使用较大比例因子复原图像时，降低视觉伪影，提高图像质量。Yang 等[96]提出深度循环融合卷积神经网络（deep recurrent fusion network，DRFN）。DRFN 采用转置卷积，避免了预处理伪影问题。采用循环残差块逐步恢复高频信息，并通过多层次特征融合充分利用不同感受野的特征，通过卷积层融合特征图重建高分辨率图像。

为了解决超分辨图像过度平滑，并丢失纹理细节的问题，Zhong 等人[97]提出超分辨率团结构网络（super-resolution clique network，SRCliqueNet）。SRCliqueNet 结合团块组和团上采样模块来从低分辨率图像中提取特征图，并利用预测小波变换粗略地重建高分辨率图像。最后，通过逆离散小波变换生成最终的高分辨率图像，从而增强纹理细节和视觉效果。考虑到超分图像的高频细节，Wang 等[98]提出端到端的深层和浅层卷积神经网络（end-to-end deep and shallow network，EEDS）。EEDS 联合训练一个深层和浅层的网络使得统一到一个框架中，浅层网络负责提取低频信息，深层网络负责捕获高频细节，两个网络相结合能同时使得增强高低频信息关联性，提高恢复图像的质量。表 7 概述了更多使用转置卷积的超分辨网络。

**表 7　基于转置卷积的单图像超分辨率卷积神经网络**

**Table 7 Single Image Super-Resolution Convolutional Neural Networks Based on Transposed Convolution**

| 作者 | 方法 | 提出时间 | 应用 | 关键词 |
|---|---|---|---|---|
| Dong 等 | FSRCNN[18] | 2016 年 | 图像超分辨 | 快速超分辨卷积神经网络 |
| Li 等 | CTU [89] | 2017 年 | 图像超分辨 | 采用帧内编码块上采样方法的卷积神经网络 |
| Mao 等 | RED-NET[90] | 2016 年 | 图像超分辨 | 深度残差的编码器-解码器卷积神经网络 |
| Lai 等 | LapSRN[91] | 2018 年 | 图像超分辨 | 拉普拉斯金字塔超分辨率卷积神经网络 |
| Hui 等 | IDN[54] | 2018 年 | 图像超分辨 | 信息蒸馏的卷积神经网络 |
| Shi 等 | FSCWRN[92] | 2019 年 | 图像超分辨 | 具有固定跳跃连接的广泛残差卷积神经网络 |
| Li 等 | SRFBN[93] | 2019 年 | 图像超分辨 | 超分辨率反馈卷积神经网络 |
| Zeng 等 | MMCN[94] | 2019 年 | 图像超分辨 | 多对多连接的卷积神经网络 |
| Haris 等 | DBPN[95] | 2018 年 | 图像超分辨 | 深度反向投影卷积神经网络 |
| Yang 等 | DRFN[96] | 2019 年 | 图像超分辨 | 深度循环融合卷积神经网络 |
| Zhong 等 | SRCliqueNet[97] | 2018 年 | 图像超分辨 | 超分辨率团结构网络 |
| Wang 等 | EEDS[98] | 2019 年 | 图像超分辨 | 端到端的深层和浅层卷积神经网络 |
| Tong 等 | SRDenseNet[21] | 2017 年 | 图像超分辨 | 使用密集跳跃连接的图像超分辨网络 |
| Li 等 | MSRN[99] | 2018 年 | 图像超分辨 | 多尺度残差的超分辨率卷积神经网络 |
| Tan 等 | SCFFN[100] | 2022 年 | 图像超分辨 | 基于自校准特征融合的高效超分辨率卷积神经网络 |
| Cheng 等 | ResLap[101] | 2020 年 | 图像超分辨 | 融合残差密集块的拉普拉斯金字塔超分辨卷积神经网络 |
| Jia 等 | MA-GAN[102] | 2022 年 | 图像超分辨 | 多头注意力生成对抗网络 |
| Chen 等 | MFFN[103] | 2024 年 | 图像超分辨 | 基于多级特征融合网络的图像超分辨率卷积神经网络 |
| Zhang 等 | UMCTN[104] | 2024 年 | 遥感图像超分辨 | 不确定性驱动的混合卷积与Transformer结合的遥感图像超分辨网络 |
| Bai 等 | BSRN[105] | 2024 年 | 图像超分辨 | 极其轻量级的图像超分辨率卷积神经网络 |
| Dargahi 等 | CPDSCNN[106] | 2023 年 | 图像超分辨 | 级联并行结构的深浅卷积神经网络 |

2）基于转置卷积的卷积神经网络图像盲超分辨方法

由于复杂的真实场景，恢复图像盲超分辨效果不佳。学者们利用生成式网络的博弈原理恢复细节信息，提高图像超分辨效果。例如，Xu 等[107]采用生成对抗网络学习特定类别的细节信息来提出了多级生成对抗网络（multi-class GAN，MCGAN）。该网络引入新的训练损失来恢复图像细节信息，并采用转置卷积来重构高质量图像。为了放大未知缩放因子的图像超分辨，Tao 等[108]提出频谱到内核的卷积神经网络（spectrum to kernel network，S2K）。S2K 充分利用低分辨率图像频谱中的形状结构，并通过可行的隐式跨领域转换直接输出相应的上采样空间核来降低了模糊核估计误差，使非盲超方法能在盲超环境下使用。更多的基于转置卷积神经网络的图像盲超分辨率方法如表 8 所示。

**表 8　基于转置卷积神经网络的图像盲超分辨方法**

**Table 8 Image Blind Super-Resolution Methods Based on Transposed Convolutional Neural Network**

| 作者 | 方法 | 提出时间 | 应用 | 关键词 |
|---|---|---|---|---|
| Xu 等 | MCGAN[107] | 2017 年 | 图像盲超分辨 | 多级生成对抗网络 |
| Liang 等 | MANet[76] | 2021 年 | 图像盲超分辨 | 相互仿射的卷积神经网络 |
| Tao 等 | S2K[108] | 2021 年 | 图像盲超分辨 | 频谱到内核的卷积神经网络 |
| Fang 等 | MAP[109] | 2022 年 | 图像盲超分辨 | 基于联合最大后验方法的卷积神经网络 |
| Du 等 | X-MDFB[110] | 2021 年 | 图像盲超分辨 | 基于多重蒸馏反馈网络的 X 射线图像超分辨率卷积神经网络 |

### 2.2.2 亚像素层

1）基于亚像素层的卷积神经网络图像超分辨方法

亚像素的核心思想是利用卷积层获得多通道数的特征图，并利用子像素卷积或其他重排列方式将多个通道的特征图在像素级上重新排列成预定尺度的高分辨率图像。由于神经网络通过卷积层可以利用更多的全局信息和上下文信息，比插值方法获得更丰富的细节信息，还原更逼真的图像信息。此外，不同于其他模块化的方法，该方法仅利用卷积和重排列操作，因此计算量小于反卷积、反池化操作，这可以大幅度提升神经网络的运行速度。基于亚像素层的上述特性， Shi 等[11]提出了高效亚像素卷积网络解决图像超分辨问题（efficient sub-pixel convolutional neural network, ESPCN）。ESPCN 在低分辨率空间中提取低分辨率特征映射，并使用亚像素层代替双三次插值操作把获得特征放大成高分辨率特征，这会降低计算成本，提高图像超分辨效率。为了进一步提高图像超分辨性能，RFB-ESRGAN[72]交替使用最近邻插值和亚像素卷积操作来充分利用输入图像的深度信息和空间信息，获得更多的细节信息，提高图像超分辨效果。为了减少图像重建过程中细节信息的丢失，Song 等[112]提出了增强深度残差的超分辨率卷积神经网络（gradual deep residual network for super-resolution, GDSR）。GDSR 使用残差块提取图像的深度特征，并将特征信息通过亚像素卷积层上采样到高分辨率空间。最后，通过多个子网络逐步重构出高质量高分辨率图像，这能有效减少了细节信息的丢失，提高图像超分辨性能为了降低图像超分辨网络的计算复杂度，Jiang 等[113]提出增强亚像素卷积神经网络（improved sub-pixel convolutional neural network, ISCNN）。该网络首先利用较小的卷积核初步地提取图像低频特征。随后，通过亚像素层上采样放大图像特征，最后利用卷积运算二次提取高频特征，以上两步操作提高了图像质量。

随着超分辨技术不断进步，对图像细节的修复

提出了更高的要求。为了有效地加强模型不同层间的信息特征相关性，Niu 等[114]提出了整体注意力卷积神经网络（holistic attention network, HAN）。该网络由层注意模块和通道－空间注意力模块组成，并通过加强模型层、通道和位置之间的整体相互依赖关系来建模，提取层次中丰富的特征，提高图像超分辨性能。为了降低计算复杂度，Jiang 等[115]提出分层密集连接的卷积神经网络（hierarchical dense connection network, HDN）。HDN 构造了分层密集残差块，并以共享的方式互相连接，允许网络在不同阶段融合特征。同时，利用分层矩阵结构提高特征表示能力，可以为信息融合和梯度优化提供额外的交错路径，但网络层数相对较少，减少了计算代价。

尽管基于深度网络的 SISR 在传统的误差度量和感知质量方面都取得了较好的效果，然而当在上采样因子较大时，这些方法仍具有挑战性。针对这一问题，Wang 等[50]提出渐进的超分辨率卷积神经网络（progressive super- resolution, ProSR）。ProSR 采用逐渐多次执行上采样操作，逐步地重建高分辨率图像。与已有的渐进式超分辨模型相比，ProSR 简化了网络内的信息传播，在上采样因子较高时保证了图像复原质量。为了解决 SISR 中任意超分辨尺度因子的问题，Hui 等[116]利用蒸馏和选择性融合的级联信息多重蒸馏块来提出信息多重蒸馏卷积神经网络（information multi-distillation network, IMDN）。其中，蒸馏模块逐步提取层次特征，融合模块根据候选特征的重要性进行聚合。最终，利用自适应剪裁策略实现任意比例因子的图像复原功能。为了降低模型的计算负担，Wa 等[117]提出了基于位移卷积的网络（shift-conv-based network, SCNet）。该网络全部采用 1×1 卷积提取图像特征来降低计算量，并通过空间位移操作扩大其感受野，在保持了超分辨性能的同时显著提高了计算效率。

然而，尽管堆叠深度小核卷积能够提高模型性能，但其仍然无法更好地扩大感受野[118]。为了解决这个问题，Wang 等[119]提出了多尺度注意力网络（multi-scale attention network, MAN），MAN 首先将图像特征通道划分为三组并在每组通道上使用不同尺寸的大核深度卷积提取多尺度特征，随后，通过结合空间注意力和门控机制调节融合的特征权重。最后，通过亚像素层上采样放大图像，实现图像的超分辨率。表 9 展示了更多基于亚像素层的图像超分辨卷积神经网络。

**表 9　基于亚像素层的单图像超分辨率卷积神经网络**

**Table 9 Single Image Super-Resolution Convolutional Neural Networks Based on Sub-Pixel Layers**

| 作者 | 方法 | 提出时间 | 应用 | 关键词 |
|---|---|---|---|---|
| Shi 等 | ESPCN[111] | 2016 年 | 图像超分辨 | 高效亚像素卷积神经网络 |
| Shang 等 | RFB-ESRGAN[72] | 2020 年 | 图像超分辨 | 基于感受野块的增强超分辨率卷积神经网络 |
| Zhao 等 | MSCNNS[120] | 2019 年 | 图像超分辨 | 多尺度亚像素卷积神经网络 |
| Song 等 | GDSR[112] | 2021 年 | 图像超分辨 | 增强深度残差的超分辨率卷积神经网络 |
| Vu 等 | FEQE[80] | 2021 年 | 图像超分辨 | 快速且高效的增强卷积神经网络 |
| Tian 等 | CFSRCNN[46] | 2020 年 | 图像超分辨 | 由粗到细卷积神经网络 |
| Jiang 等 | ISCNN[113] | 2021 年 | 图像超分辨 | 增强亚像素卷积神经网络 |
| Niu 等 | HAN[114] | 2020 年 | 图像超分辨 | 整体注意力卷积神经网络 |
| Jiang 等 | HDN[115] | 2020 年 | 图像超分辨 | 分层密集连接的卷积神经网络 |
| Wang 等 | ProSR[50] | 2018 年 | 图像超分辨 | 渐进结构和训练的超分辨率卷积神经网络 |
| Hui 等 | IMDN[116] | 2019 年 | 图像超分辨 | 信息多重蒸馏卷积神经网络 |
| Ruan 等 | ESCNN[121] | 2022 年 | 图像超分辨 | 高效亚像素卷积神经网络 |
| Xie 等 | LKDN[122] | 2023 年 | 图像超分辨 | 大核蒸馏卷积神经网络 |
| Fang 等 | HNCT[123] | 2022 年 | 图像超分辨 | 卷积与注意力融合网络 |
| Kong 等 | RLFN[124] | 2022 年 | 图像超分辨 | 残差局部特征卷积神经网络 |
| Tian 等 | DSRNet[125] | 2023 年 | 图像超分辨 | 动态的超分辨率卷积神经网络 |

2）基于亚像素层的卷积神经网络图像盲超分辨方法

基于核估计的盲超分辨率方法，在复杂的退化情况下难以准确估计核，导致超分辨效果不佳。为了提高核估计精准性，Xiao 等[126]提出了渐进式盲超分辨率卷积神经网络（Progressive CNN model for blind Super-Resolution, PCSR）。PCSR 直接从概率图模型的角度出发，使用密集连接和渐进策略有效地利用跨尺度的图像先验信息，重建高质量图像。由于真实图像往往保有随机模糊和噪声，这会增加了

图像超分辨难度。为解决这一问题，Huo 等[127]提出空间上下文幻觉的卷积神经网络（spatial context hallucination network, SCHN），它能将去噪、去模糊和超分辨集成到一个框架中。SCHN 利用可变形卷积纠正由于卷积操作引起的图像像素偏移问题，并利用像素重组卷积扩大空间维度，避免模糊核未知时，积累的误差会影响图像去模糊和超分辨率效果，从而降低复原图像的质量。为了解决上述问题，Dong 等[128]提出基于模糊核预测方法的卷积神经网络（super-resolution method with blur kernel prediction, BKPSR）。BKPSR 使用轻量级的卷积神经网络和变分自编码器来预测模糊核代码，随后利用预测的模糊核代码辅助图像超分辨，取得了很好的效果。表 10 给出了更多基于亚像素层的单图像盲超分辨率卷积神经网络。

表 10 基于亚像素层的单图像盲超分辨率卷积神经网络

**Table 10 Single Image Blind Super-Resolution Convolutional Neural Networks Based on Sub-Pixel Layers**

| 作者 | 方法 | 提出时间 | 应用 | 关键词 |
|---|---|---|---|---|
| Xiao 等 | PCSR[126] | 2019 年 | 图像盲超分辨 | 多重退化盲超分辨的深度渐进卷积神经网络 |
| Huo 等 | SCHN[127] | 2020 年 | 图像盲超分辨 | 空间上下文幻觉的卷积神经网络 |
| Dong 等 | BKPSR[128] | 2021 年 | 图像盲超分辨 | 基于模糊核预测方法的卷积神经网络 |

2.2.3 元上采样模块

1）基于元上采样模块的卷积神经网络图像超分辨方法

在真实世界中，退化的分辨率未知，因此可根据使用者的需求任意倍数恢复图像的分辨率尤为重要[129]。传统的机器学习方法针对不同因子训练多个上采样模块，来预定义缩放因子，效率较低。而基于元上采样模块是基于元学习的思想，该方法的特点在于不再局限于固定的放大倍数，而是通过一个灵活的模块动态生成适配于特定倍数的滤波器参数，从而提升了在任意放大倍数下的适用性。具体而言，这类方法通过网络来预测滤波器参数，使模型能够在不同的上采样因子条件下对图像进行有效的超分辨处理，从而提高了计算效率和适应性。不同于之前方法需要根据不同放大倍数设计不同的插值参数和模型结构，这种基于预测的参数生成方式可以在不牺牲性能的情况下简化模型的结构，使得不同的放大倍数可以使用相同的模型结构。Hu 等[129]首次利用单一模型求解任意尺度因子的图像超分辨率，并提出了元超分辨率卷积神经网络（meta-super resolution, Meta-SR）。对于不同的缩放因子，Meta-SR 使用元上采样模块代替传统模块，元上采样模块动态预测上采样过滤器权重，使其能够在不同尺度因子下工作，通过特征图与过滤器间的卷积计算，生成任意大小的高分辨率图像。由于多数超分辨网络仅关注固定整数尺度因子的图像 SR，并仅能在单一模型中处理如采样因子、模糊核和噪声水平中一种退化参数，而导致实际使用效率不高。

为解决上述问题，Hu 等[130]基于 Meta-SR 的基础上，提出了元统一超分辨率卷积神经网络（meta-unified super-resolution network, Meta-USR）。该网络能利用元修复块增强传统的上采样模块，自适应预测各种退化参数组合的卷积滤波器权值，解决复杂的退化参数的图像超分辨问题。为了实现精确、灵活的任意尺度因子 SISR，Fu 等[131]提出残差尺度注意力卷积网络（residual scale attention network, RSAN）。RSAN 将尺度因子作为先验知识引入深度网络模型，学习低分辨率图像的判别特征，并利用坐标信息和尺度因子的二次多项式预测像素级重建核，实现任意尺度因子的图像超分辨率。为了充分利用图像潜在的多尺度特征，Liu 等[99]提出多尺度跳跃连接的卷积神经网络（multi-scale skip-connection network, MSN）。MSN 首先利用不同大小的卷积核从低分辨率图像中提取多尺度特征。然后，将这些特征输入多尺度混合组以充分提取图像细节信息。随后，通过混合卷积层训练来自前一尺度和当前尺度的图像细节信息，每层输出通过跳跃连接传递到后续混合卷积层，形成密集连接；最后，通过充分利用图像的潜在多尺度特征，采用元上采样模块实现任意比例因子的特征图放大，从而重建高分辨率图像。

由于元上采样模块在图像处理方面的优越性，不同应用领域纷纷使用该模块提高图像超分质量与灵活性，例如，在光图像融合领域，Li 等[132]提出基于元学习的红外与可见光图像融合深度框架。在该框架中，首先通过卷积神经网络提取源图像特征并根据使用者的需求，使用任意比例因子的元上采样模块得到高分辨率图像特征；随后，输入基于双重注意力机制的特征融合模块，实现不同源图像的特征融合；最后，在框架中迭代使用残差补偿模块，增强网络的细节提取能力，提高光图像融合效果。在医学成像领域，Tan 等[133]提出元超分辨生成对抗网络（Meta-SRGAN），该网络在 SRGAN 的基

础上，利用元上采样模块实现任意尺度、高保真的大脑核磁共振图像超分辨率，辅助医疗诊断。Zhu 等[134]提出任意尺度医学图像超分辨率卷积神经网络（Medical Image Arbitrary Scale Super-Resolution，MIASSR），该方法将元学习与生成对抗网络相结合，并利用迁移学习实现全新医疗模式的超分辨任务，如心脏磁共振图像和胸部计算机断层扫描图像超分辨任务。表 11 概括了更多的基于元上采样模块卷积神经网络图像超分辨方法。

**表 11 基于元上采样模块的卷积神经网络图像超分辨方法**

**Table 11 Convolutional Neural Network Image Super-Resolution Methods Based on Meta-Upsampling Modules**

| 作者 | 方法 | 提出时间 | 应用 | 关键词 |
|---|---|---|---|---|
| Hu 等 | Meta-SR[129] | 2019 年 | 图像超分辨 | 元超分辨率卷积神经网络 |
| Hu 等 | Meta-USR[130] | 2020 年 | 图像超分辨 | 元统一超分辨率卷积神经网络 |
| Fu 等 | RSAN[131] | 2021 年 | 图像超分辨 | 残差尺度注意力卷积网络 |
| Tan 等 | Meta-SRGAN[133] | 2020 年 | 图像超分辨 | 元超分辨生成对抗网络 |
| Zhu 等 | MIASSR[134] | 2021 年 | 图像超分辨 | 任意尺度医学图像超分辨率卷积神经网络 |
| Liu 等 | MSN[99] | 2018 年 | 图像超分辨 | 多尺度跳跃连接的卷积神经网络 |
| Hong 等 | DC-Net[135] | 2023 年 | 图像超分辨 | 解耦与耦合的卷积神经网络 |

2）基于元上采样模块的卷积神经网络图像盲超分辨方法

为提高真实图像的盲超分辨效果，一些方法试图用多种退化因子的复杂组合训练超分辨网络，以覆盖真实的退化空间。但由于运动的物体和失焦等因素，模糊核的估计误差会导致图像超分辨失效。针对这一问题，Xia 等[136]提出基于元学习的区域退化感知超分辨率卷积神经网络（Meta-Learning based Region Degradation Aware SR Network, MRDA）。MRDA 包括三种不同网络类型：元学习网络、退化提取网络和区域退化感知超分辨网络。首先，轻量级元学习网络采用元学习算法学习退化过程；随后，退化提取网络通过多次迭代，快速适应特定的复杂退化，并隐式提取退化信息；最后，区域退化感知超分辨网络利用空间调制系数，自适应调整显式退化表示的影响，提高图像盲超分辨率性能。

## 3 性能比较

为了使读者更方便地了解到更多面向图像超分辨的卷积神经网络方法，本节首先介绍不同方法常用数据集、常用评价指标、实验设置，然后将从定量比较和可视化分析两个方面比较不同图像超分辨方法的性能与表现，更多的信息如下所示：

### 3.1 数据集

本小节对上述的图像超分辨模型划分为两类:非盲图像超分辨率卷积神经网络与盲图像超分辨率卷积神经网络。数据集通常分为训练数据集和测试数据集，分别用于模型的训练与评估。此外，验证数据集通常包含在训练数据集中，用来训练过程中检测图像超分辨模型效果，它通常是与测试集具有相同分布的图像组成的数据集。不同模型的更多细节内容如下所示：

1）单图像超分辨率卷积神经网络

训练数据集：ImageNet[137]，91-images[7]，BSD[138]，DIV2K[139]，Harvard[140]，General-100[141]，Flickr2K[139]，MSCOCO[142]，OST[143]，DIV8K[144]，SRResNet[25]，WED[145]，FFHQ[146]，Nico-illust[147]，NYU Depth[148]，Make3D[149]，T91[7]，OASIS[150]，BraTS[151]，ACDC[152]，COVID-CT[153]，ILSVRC 2012[137]，UCID[154]。

测试数据集：Set5[155]，Set14[156]，BSD，Urban100[141]，Manga109[157]，DIV2K[139]，CAVE[158]，Harvard，720p[159]，PIRM[160]，DIV8K，Nico-illust，SRResNet，DIV2K4D[161]，Set12[162]，UCID，SuperTexture[163]，NYU Depth，Make3D，ImageNet400[137]，OASIS，BraTS，ACDC，COVID-CT。

2）单图像盲超分辨率卷积神经网络

训练数据集：DIV2K，Flickr2K，WED，FFHQ，OutdoorSceneTraining[143]，DIV2KRK[60]，CelebA[164]，VOC2012[165]。

测试数据集：NTIRE2018[166]，Set5，Set14，BSD，DIV2K，Urban100，RealSRSet[161]，OST，DPED[167]，ADE20K[168]，RealSR[169]，DRealSR[170]，CelebA，Manga109。

为了使读者更方便地了解到相关图像超分辨模型的数据集信息，表 12 概括了上述不同单图像超分辨率和盲超分辨率方法的训练数据集和测试数据集。

**表 12 单图像超分辨率与盲超分辨率的训练与测试数据集**

**Table 12 Training and testing datasets for single image super-resolution and blind super-resolution**

| 方法分类 | 图像超分辨方法 | 训练数据集 | 测试数据集 |
|---|---|---|---|
| 单图像超分辨率 | SRCNN[16] | ImageNet | Set5, Set14 |
| | VDSR[17] | T91, BSD | Set5, Set14, BSD100, Urban100 |
| | CMSC[40] | T91, BSD300 | Set5, Set14, BSD100, Urban100 |
| | DRCN[20] | T91 | Set5, Set14, BSD100, Urban100 |
| | DRRN[43] | T91, BSD | Set5, Set14, BSD100, Urban100 |
| | MemNet[39] | BSD | Set5, Set14, BSD100, Urban100 |
| | EDSR[44] | DIV2K | Set5, Set14, BSD100, Urban100, DIV2K |
| | MDSR[44] | DIV2K | Set5, Set14, BSD100, Urban100, DIV2K |
| | ProSR[50] | DIV2K | Set5, Set14, BSD100, Urban100, DIV2K |
| | RCAN[23] | DIV2K | Set5, Set14, BSD100, Urban100, Manga109 |
| | DCSCN[41] | T91, BSD | Set5, Set14, BSD100 |
| | SSF-CNN[52] | Harvard | CAVE, Harvard |
| | LISTA[53] | T91 | Set5, Set14, BSD100 |
| | IDN[54] | T91, BSD | Set5, Set14, BSD100, Urban100 |
| | CFSRCNN[46] | DIV2K | Set5, Set14, BSD100, Urban100, 720p |
| | FSRCNN[18] | T91, General-100 | Set5, Set14, BSD100 |
| | USRNet[69] | DIV2K, Flickr2K | BSD68 |
| | BT-SRN[70] | DIV2K | DIV2K |
| | EnhanceNet[70] | MSCOCO | Set5, Set14, BSD100, Urban100 |
| | ESRGAN[26] | DIV2K, Flickr2K, OST | Set5, Set14, BSD100, Urban100, PIRM |
| | RFB-ESRGAN[72] | DIV8K, DIV2K, Flickr2K, OST | DIV8K |
| | PAN[73] | DIV2K, Flickr2K | Set5, Set14, BSD100, Urban100, Manga109 |
| | CCNet[74] | SRResNet | SRResNet |
| | Blind ESRGAN[62] | DIV2K, Flickr2K, WED, FFHQ | DIV2K4D, RealSRSet |
| | MANet[76] | DIV2K | Set5, Set14, BSD100, Urban100 |
| | MC-SRCNN[78] | T91 | Set5, Set14 |
| | m-SRCNN[79] | Nico-illust | Nico-illust |
| | FEQE[80] | DIV2K | Set5, Set14, BSD100, Urban100 |
| | PRLSR[81] | DIV2K | Set5, Set14, BSD100, Urban100 |
| | SCN[84] | T91 | Set12, BSD64, Urban100 |
| | DefRCN[82] | DIV2K | Set5, Set14, BSD100, Urban100 |
| | CTU[89] | UCID | UCID |
| | RED-NET[90] | BSD | BSD200 |
| | LapSRN[91] | T91, BSD200 | Set5, Set14 |
| | IDN[54] | T91, BSD200 | Set5, Set14, BSD100, Urban100 |
| | SRFBN[93] | DIV2K, Flickr2K | Set5, Set14, BSD100, Urban100, Manga109 |
| | MMCN[94] | T91 | Set5, Set14, BSD100, Urban100 |
| | DBPN[95] | DIV2K, Flickr2K, ImageNet | Set5, Set14, BSD100, Urban100, Manga109 |
| | DRFN[96] | T91, BSD200 | Set5, Set14, BSD100, Urban100, ImageNet400 |
| | SRCliqueNet[97] | DIV2K | Set14, BSD100, Urban100 |
| | EEDS[98] | T91 | Set5, Set14, BSD100 |
| | SRDenseNet[21] | ImageNet | Set5, Set14, BSD100, Urban100 |
| | MSRN[99] | DIV2K | Set5, Set14, BSD100, Urban100, Manga109 |
| | ESPCN[111] | ImageNet | Set5, Set14, BSD300, BSD500, SuperTexture |
| | MSCNNS[120] | NYU Depth, Make3D | NYU Depth, Make3D |
| | GDSR[112] | T91, BSD200 | Set5, Set14, BSD100, Urban100 |
| | ISCNN[113] | BSD300 | BSD300, BSD500 |
| | HAN[114] | DIV2K | Set5, Set14, BSD100, Urban100, Manga109 |
| | HDN[115] | DIV2K | Set5, Set14, BSD100, Urban100, Manga109 |
| | IMDN[116] | DIV2K | Set5, Set14, BSD100, Urban100, Manga109 |
| | Meta-RDN[129] | DIV2K | Set14, BSD100, Manga109, DIV2K |
| | Meta-USR[130] | DIV2K | DIV2K, Set5, Set14, BSD100, Urban100, Manga109 |
| | RSAN[131] | DIV2K | Set5, Set14, BSD100, Urban100 |
| | Meta-SRGAN[133] | DIV2K, BraTS | BraTS |
| | MIASSR[134] | OASIS, BraTS, ACDC, COVID-CT | OASIS, BraTS, ACDC, COVID-CT |
| | MSN[99] | ILSVRC 2012, DIV2K | DIV2K, Set5, Set14, BSD100, Urban100, Manga109 |
| 单图像盲超分辨率 | CinCGAN[55] | DIV2K | NTIRE2018 |
| | UISR[56] | DIV2K | NTIRE2018 |
| | IKC[58] | DIV2K, Flickr2K | Set5, Set14, BSD100 |
| | KMSR[59] | DIV2K | DIV2K |
| | KernelGAN[60] | - | NTIRE2018 |
| | DASR[61] | DIV2K, Flickr2K | Set5, Set14, BSD100, Urban100 |
| | BlindESRGAN[62] | DIV2K, Flickr2K, WED, FFHQ | DIV2K, RealSRSet |
| | RealESRGAN[45] | DIV2K, Flickr2K, OutdoorSceneTraining | RealSR, DRealSR, OST300, DPED, ADE20K |
| | EDKE[87] | DIV2KRK | RealSR |
| | MCGAN[107] | CelebA | CelebA |
| | MANet[76] | DIV2K | BSD100 |
| | S2K[108] | DIV2K | DIV2K, Flickr2K |
| | PCSR[126] | DIV2K, VOC2012 | Set5, Set14 |
| | SCHN[127] | DIV2K, Flickr2K | Set5, Set14 |
| | MRDA[136] | DIV2K, Flickr2K | Set5, Set14, BSD100, Urban100, Manga109 |

### 3.2 评价指标

为定量描述不同方法的性能差异，本节将介绍常用于图像超分辨的评价指标，包括峰值信噪比（Peak Signal-to-Noise Ratio, PSNR）[171]、结构相似性（Structural Similarity, SSIM）[172]、模型参数量以及运行时间。

峰值信噪比（Peak Signal-to-Noise Ratio, PSNR）和结构相似性（Structural Similarity, SSIM）是两个著名的客观图像质量评价指标[171]。PSNR 用于衡量重建图像与参考图像之间的相似度，通过计算图像之间像素的均方误差（Mean Square Error, MSE）来实现。PSNR 的计算公式如下：

$$M_{PSNR}=10\cdot\log_{10}(\frac{P_{MAX}}{M_{MSE}})^2 \quad (1)$$

其中，$M_{PSNR}$ 表示计算得到峰值信噪比的值。$P_{MAX}$ 表示图像中可取最大的像素值，一般为 255。$M_{MSE}$ 表示重建图像与参考图像之间像素值的均方误差，计算公式如下：

$$M_{MSE}=\frac{1}{m\times n}\sum_{i=1}^{m}\sum_{j=1}^{n}(I_{SR}^{i,j}-I_{HR}^{i,j})^2 \quad (2)$$

其中，$M_{MSE}$ 表示计算得到的均方误差值，$I_{SR}^{i,j}$ 代表重建图片的第 $i$ 行第 $j$ 列的像素值，$I_{HR}^{i,j}$ 代表参考图片的第 $i$ 行第 $j$ 列的像素值。$m$ 和 $n$ 分别表示图像的行和列大小。

SSIM 用于模拟人类视觉对图像结构信息的感知[172]。与 PSNR 相比，SSIM 更加关注图像之间亮度、对比度和结构等信息的相似性。SSIM 的计算公式如下：

$$M_{SSIM}=\frac{(2u_{SR}u_{HR}+\varepsilon_1)(2v_{SR,HR}+\varepsilon_2)}{(u_{SR}^2+u_{HR}^2+\varepsilon_1)(v_{SR}^2+v_{HR}^2+\varepsilon_2)} \quad (3)$$

其中，$M_{SSIM}$ 表示计算得到结构相似性的值。$u_{SR}$ 和 $u_{HR}$ 分别表示重建图像和参考图像的平均亮度。$v_{SR}$ 和 $v_{HR}$ 分别表示重建图像和参考图像的亮度方差。$v_{SR,HR}$ 是两张图像的协方差。$\varepsilon_1$ 和 $\varepsilon_2$ 是一个极小的常数。

模型参数量和运行时间是评估模型性能和效率的两个重要指标。模型参数量用于评价模型的复杂度和内存占用，通过计算各层的权重矩阵的大小求和得到。运行时间用于评价模型在推理阶段对图片执行运算的快慢，通过计算在相同的硬件条件下对相同图片完成推理过程所需的平均时间来衡量。

### 3.3 实验设置

在图像超分辨领域，为了保持公平，测试时用 Y 通道测试[16,18]。由于不同方法实验设备、配置及实验设置不同，通常选择 PSNR 和 SSIM 及可视化图像比较细节来验证获得图像超分辨模型的性能。每种方法的实验设置可参考具体方法论文中参数设置。此外，待恢复倍数代表将低分辨率图像放大到几倍高清图像。

### 3.4 定量比较

为验证第 3 章中提到的图像超分辨方法性能，本节将比较在非盲超分辨率与盲超分辨率两类方法在公开数据集 上，通过测试这些方法的峰值信噪比、结构相似性、模型参数量以及运行时间，对这些方法进行定量分析。

1）单图像超分辨模型的定量分析

本次测试了 VDSR、CMSC、DRCN、DRRN、MemNet、EDSR、MDSR、RCAN、LISTA、IDN、CFSRCNN、FSRCNN、LapSRN、MANet、CNF、DefRCN、SRFBN、MSRN、HAN、RDN 等方法在 Set5、Set14、BSD100、Urban100 等公共数据集上

不同倍数的图像超分辨性能。在表 13 中，在 Set5 数据集上倍数为 2 时，RDN 的性能表现最好，在表 14 中倍数为 3 和表 15 中 4 时，RCAN 的性能表现最好。此外，本实验选取部分超分辨模型在不同大小的图像上实际运行，比较不同模型在×2 放大倍率时的处理时间。表 16 列举出的多种模型当中，CFSRCNN 的模型处理速度最快。最后，本实验比较了几种经典单图像超分辨模型的参数大小与计算速度，如表 17 所示，CARN-M[173]的参数规模最小、运算速度最快、模型复杂度最小。

2）单图像盲超分辨率模型定量分析

对于众多盲超分辨率模型，本实验同样选取了 Set5 和 Set14 公共数据集进行实验，表 18 列举了经典单图像盲超分辨模型在常见数据集、不同倍率时的性能表现，每种超分辨方法对应右侧第一行数值为 PSNR，第二行数值为 SSIM。如表 18 所示，在 Set5 数据集上且倍率为 2 和 4 时，DASR 性能表现最好。

表 13　单图像超分辨率模型在 Set5、Set14、BSD100、Urban100 上×2 倍率时 PSNR（dB）与 SSIM

**Table 13 The evaluation of PSNR (dB) and SSIM Results for scale factor 2 in Single Image Super-Resolution on Set5, Set14, BSD100, and Urban100**

| 方法 | 放大倍数 | Set5 PSNR/SSIM | Set14 PSNR/SSIM | BSD100 PSNR/SSIM | Urban100 PSNR/SSIM |
|---|---|---|---|---|---|
| VDSR[17] | ×2 | 35.53/0.9587 | 33.03/0.9124 | 31.90/0.8960 | 30.76/0.9140 |
| CMSC[40] | | 37.89/0.9605 | 33.41/0.9153 | 32.15/0.8992 | 31.47/0.9220 |
| DRCN[20] | | 37.63/0.8588 | 33.06/0.9121 | 31.85/0.8942 | 30.76/0.9133 |
| DRRN[43] | | 37.74/0.9591 | 33.23/0.9136 | 32.05/0.8973 | 31.23/0.9188 |
| MemNet[39] | | 37.68/0.9597 | 33.28/0.9142 | 32.08/0.8987 | 31.31/0.9195 |
| EDSR[44] | | 38.02/0.9606 | 34.02/0.92.4 | 32.37/0.9018 | 33.10/0.9363 |
| MDSR[44] | | 38.17/0.9605 | 33.92/0.9203 | 32.34/0.9014 | 33.03/0.9362 |
| RCAN[23] | | 38.33/0.9617 | 34.23/0.9225 | 32.46/0.9031 | 33.54/0.9399 |
| LISTA[53] | | 37.41/0.9567 | 32.71/0.9095 | 31.54/0.8908 | - |
| IDN[54] | | 37.83/0.9600 | 33.30/0.9148 | 32.08/0.8985 | 31.27/0.9196 |
| CFSRCNN[46] | | 37.79/0.9591 | 33.51/0.9165 | 32.11/0.8988 | 32.07/0.9273 |
| FSRCNN[18] | | 36.98/0.9556 | 32.62/0.9087 | 31.50/0.8904 | 29.85/0.9009 |
| LapSRN[91] | | 37.52/0.9591 | 33.08/0.9130 | 31.08/0.8950 | 30.41/0.9101 |
| MANet[76] | | 35.98/0.9420 | 31.95/0.8845 | 30.97/0.8650 | 29.87/0.8877 |
| CNF[38] | | 37.66/0.9690 | 33.08/0.9136 | 31.91/0.8962 | - |
| DefRCN[82] | | 38.02/0.9596 | 33.58/0.9151 | 32.21/0.8998 | 32.20/0.9286 |
| SRFBN[93] | | 38.11/0.9609 | 33.82/0.9196 | 32.29/0.9010 | 32.62/0.9328 |
| MSRN[99] | | 38.08/0.9605 | 33.74/0.9170 | 32.23/0.9013 | 32.22/0.9326 |
| HAN[114] | | 26.83/0.7919 | 23.21/0.6888 | 25.11/0.6613 | 22.42/0.6571 |
| RDN[22] | | 38.24/0.9614 | 34.01/0.9212 | 32.34/0.9017 | 32.89/0.9353 |

表 14　单图像超分辨率模型在 Set5、Set14、BSD100、Urban100 上×3 倍时 PSNR（dB）与 SSIM

**Table 14 The evaluation of PSNR (dB) and SSIM Results for scale factor 3 in Single Image Super-Resolution on Set5, Set14, BSD100, and Urban100**

| 方法 | 放大倍数 | Set5 PSNR/SSIM | Set14 PSNR/SSIM | BSD100 PSNR/SSIM | Urban100 PSNR/SSIM |
|---|---|---|---|---|---|
| VDSR[17] | ×3 | 33.66/0.9213 | 29.77/0.8314 | 28.82/0.7976 | 27.14/0.8279 |
| CMSC[40] | | 34.24/0.9266 | 30.09/0.8371 | 29.01/0.8024 | 27.69/0.8411 |
| DRCN[20] | | 33.82/0.9226 | 29.77/0.8314 | 28.80/0.7963 | 27.15/0.8277 |
| DRRN[43] | | 34.03/0.9244 | 29.96/0.8349 | 28.95/0.8004 | 27.53/0.8377 |
| MemNet[39] | | 34.09/0.9248 | 30.00/0.8350 | 28.96/0.8001 | 27.56/0.8376 |
| EDSR[44] | | 34.76/0.9290 | 30.66/0.8481 | 29.32/0.8104 | 29.02/0.8685 |
| MDSR[44] | | 34.77/0.9288 | 30.53/0.8465 | 29.30/0.8101 | 28.99/0.8683 |
| RCAN[23] | | 34.85/0.9305 | 30.76/0.8494 | 29.39/0.8122 | 29.31/0.8736 |
| LISTA[53] | | 33.26/0.9167 | 29.55/0.8271 | 28.58/0.7910 | - |
| CFSRCNN[46] | | 34.24/0.9256 | 30.27/0.8410 | 29.03/0.8035 | 28.04/0.8496 |
| FSRCNN[18] | | 33.16/0.9140 | 29.42/0.8242 | 28.52/0.7893 | 26.41/0.8064 |
| LapSRN[91] | | 33.82/0.9227 | 29.87/0.8320 | 28.82/0.7980 | 27.07/0.8280 |
| MANet[76] | | 33.69/0.9184 | 29.81/0.8270 | 28.80/0.7931 | 27.39/0.8331 |
| CNF[38] | | 33.74/0.9226 | 29.90/0.8322 | 28.82/0.7980 | - |
| DefRCN[82] | | 34.41/0.9263 | 30.34/0.8388 | 29.01/0.8044 | 28.16/0.8519 |
| SRFBN[93] | | 34.70/0.9292 | 30.51/0.8461 | 29.24/0.8084 | 28.73/0.8641 |
| MSRN[99] | | 34.38/0.9262 | 30.34/0.8395 | 29.08/0.8041 | 28.08/0.8554 |
| HAN[114] | | 23.71/0.6171 | 22.31/0.5878 | 23.21/0.5653 | 20.34/0.5311 |
| RDN[22] | | 34.71/0.9296 | 30.57/0.8468 | 29.26/0.8093 | 28.80/0.8653 |
| IDN[54] | | 34.11/0.9253 | 29.99/0.8354 | 28.95/0.8013 | 27.42/0.8359 |

表 15　单图像超分辨率模型在 Set5、Set14、BSD100、Urban100 上×4 倍率时 PSNR（dB）与 SSIM

**Table 15 The evaluation of PSNR (dB) and SSIM Results for scale factor 4 in Single Image Super-Resolution on Set5, Set14, BSD100, and Urban100**

| 方法 | 放大倍数 | Set5 PSNR/SSIM | Set14 PSNR/SSIM | BSD100 PSNR/SSIM | Urban100 PSNR/SSIM |
|---|---|---|---|---|---|
| VDSR[17] | ×4 | 31.35/0.8838 | 28.01/0.7674 | 27.29/0.7251 | 25.18/0.7524 |
| CMSC[40] | | 31.91/0.8923 | 28.35/0.7751 | 27.46/0.7308 | 25.64/0.7692 |
| DRCN[20] | | 31.53/0.8854 | 28.03/0.7673 | 27.14/0.7233 | 25.14/0.7511 |
| DRRN[43] | | 31.68/0.8888 | 28.21/0.7720 | 27.38/0.7284 | 25.44/0.7638 |
| MemNet[39] | | 31.74/0.8893 | 28.26/0.7723 | 27.40/0.7281 | 25.50/0.7630 |
| EDSR[44] | | 32.62/0.8984 | 28.94/0.7901 | 27.79/0.7437 | 26.86/0.8080 |
| MDSR[44] | | 32.60/0.8982 | 28.82/0.7876 | 27.78/0.7425 | 26.86/0.8082 |
| RCAN[23] | | 32.73/0.9013 | 28.98/0.7910 | 27.85/0.7455 | 27.10/0.8142 |
| LISTA[53] | | 31.04/0.8775 | 27.76/0.7620 | 27.11/0.7191 | - |
| CFSRCNN[46] | | 32.06/0.8920 | 28.57/0.7800 | 27.53/0.7333 | 26.03/0.7824 |
| FSRCNN[18] | | 30.70/0.8657 | 27.59/0.7535 | 26.96/0.7128 | 24.60/0.7258 |
| LapSRN[91] | | 31.54/0.8850 | 28.19/0.7720 | 27.32/0.7270 | 25.21/0.7560 |
| MANet[76] | | 31.54/0.8876 | 28.28/0.7727 | 27.35/0.7305 | 25.66/0.7759 |
| CNF[38] | | 31.55/0.8856 | 28.15/0.7680 | 27.32/0.7253 | - |
| DefRCN[82] | | 32.21/0.8936 | 28.59/0.7810 | 27.57/0.7356 | 26.04/0.7841 |
| SRFBN[93] | | 32.47/0.8983 | 28.81/0.7868 | 27.72/0.7409 | 26.60/0.8015 |
| MSRN[99] | | 32.07/0.8903 | 28.60/0.7751 | 27.52/0.7273 | 26.04/0.7896 |
| HAN[114] | | 21.71/0.5941 | 20.42/0.4937 | 21.48/0.4901 | 19.01/0.4676 |
| RDN[22] | | 32.47/0.8990 | 28.81/0.7871 | 27.72/0.7419 | 26.61/0.8028 |
| IDN[54] | | 31.82/0.8903 | 28.25/0.7730 | 27.41/0.7297 | 25.41/0.7632 |

表 16　单图像超分辨模型在 256×256、512×512 和 1024×1024 大小的图像和放大倍率为×2 时的运行时间（秒）

**Table 16 Runtime (s) for Single Image Super-Resolution Models on Images of Size 256×256, 512×512, and 1024×1024 with scale factor of 2**

| 方法 \ 图像大小 | 256×256 | 512×512 | 1024×1024 |
|---|---|---|---|
| VDSR[17] | 0.0172 | 0.0575 | 0.2126 |
| DRRN[43] | 3.063 | 8.050 | 25.23 |
| MemNet[39] | 0.8774 | 3.605 | 14.69 |
| RDN[22] | 0.0553 | 0.2232 | 0.9124 |
| SRFBN[93] | 0.0761 | 0.2508 | 0.9787 |
| CARN-M[173] | 0.0159 | 0.0199 | 0.0320 |
| CFSRCNN[46] | 0.0153 | 0.0184 | 0.0298 |

表 17　单图像超分辨模型的复杂度

**Table 17 Complexity of Single Image Super-Resolution Models**

| 超分辨方法 | 参数量 | 每秒浮点运算次数 |
|---|---|---|
| VDSR[17] | 665K | 15.82G |
| DRRN[43] | 1774K | 42.07G |
| MemNet[39] | 677K | 16.06G |
| CARN-M[173] | 412K | 2.50G |
| CARN[173] | 1592K | 10.13G |
| RDN[22] | 21937K | 130.75G |
| CFSRCNN[46] | 1200K | 11.08G |
| SRFBN[93] | 3631K | 22.24G |

表 18　图像盲超分辨率模型在 Set5、Set14 数据集上×2、×3 与×4 放大倍率时 PSNR 与 SSIM

**Table 18 PSNR and SSIM for Blind Super-Resolution Models on Set5 and Set14 Datasets at ×2, ×3, and ×4 Magnifications**

| 数据集 | Set5 | Set14 | Set5 | Set14 | Set5 | Set14 |
|---|---|---|---|---|---|---|
| 方法 \ 放大倍率 | ×2 | | ×3 | | ×4 | |
| IKC[48] | 36.62<br>0.9658 | 32.82<br>0.8999 | 32.16<br>0.9420 | 29.46<br>0.8229 | 31.52<br>0.9278 | 28.26<br>0.7688 |
| DASR[61] | 37.87<br>- | 33.34<br>- | 34.11<br>- | 30.13<br>- | 31.99<br>- | 28.50<br>- |
| PCSR[126] | 31.49<br>0.909 | 29.88<br>0.856 | -<br>- | -<br>- | 29.20<br>0.807 | 26.82<br>0.694 |
| SCHN[127] | 33.70<br>0.9310 | 29.64<br>0.8689 | -<br>- | -<br>- | 27.81<br>0.8339 | 24.70<br>0.7056 |

### 3.5 可视化图像

为了使读者更方便地了解到不同单图像非盲超分辨率和盲超分辨方法的可视化效果，本节将展示部分超分辨模型的人眼观察到的结果，即不同超分辨方法的效果对比图。图 11-14 中由(a)到(h)分别为高清原始图像、传统双三次插值后的图像、SRCNN 处理图像、VDSR 处理图像、DRCN 处理图像、CARN-M 处理图像、LESRCNN[174]处理图像、CFSRCNN 处理图像。从图 11 中可以看到，在 BSD100 数据集上，放大倍数为 3 时，CFSRCNN 方法效果最好。对比其他方法，CFSRCNN 方法恢复图像的窗户框更平直，并且能恢复部分小窗户框。图 12 中，在 BSD100 数据集上，放大倍数为 4 时，CFSRCNN 方法表现最为优异，大多数方法无法恢复出游艇上最右侧的竖杆，CFSRCNN 方法可以看清有竖杆。图 13 中，在 Urban100 数据集上，放大倍数为 3 时，CFSRCNN 方法再次展示出色的性能，在更高楼层仍然保持清晰横向玻璃框，而其他方法则会变得模糊或者错误地变成竖向玻璃框。在图 14 中可以看出，在 Urban100 数据集上，放大倍数为 4 时, CFSRCNN 方法同样表现突出，能够清晰地恢复出第三条柱子，而其他方法会模糊或者缺失第三根柱子。综合来看，CFSRCNN 方法在不同数据集和放大倍数下普遍表现优越，显示出其在多种场景中卓越的图像超分辨率能力。

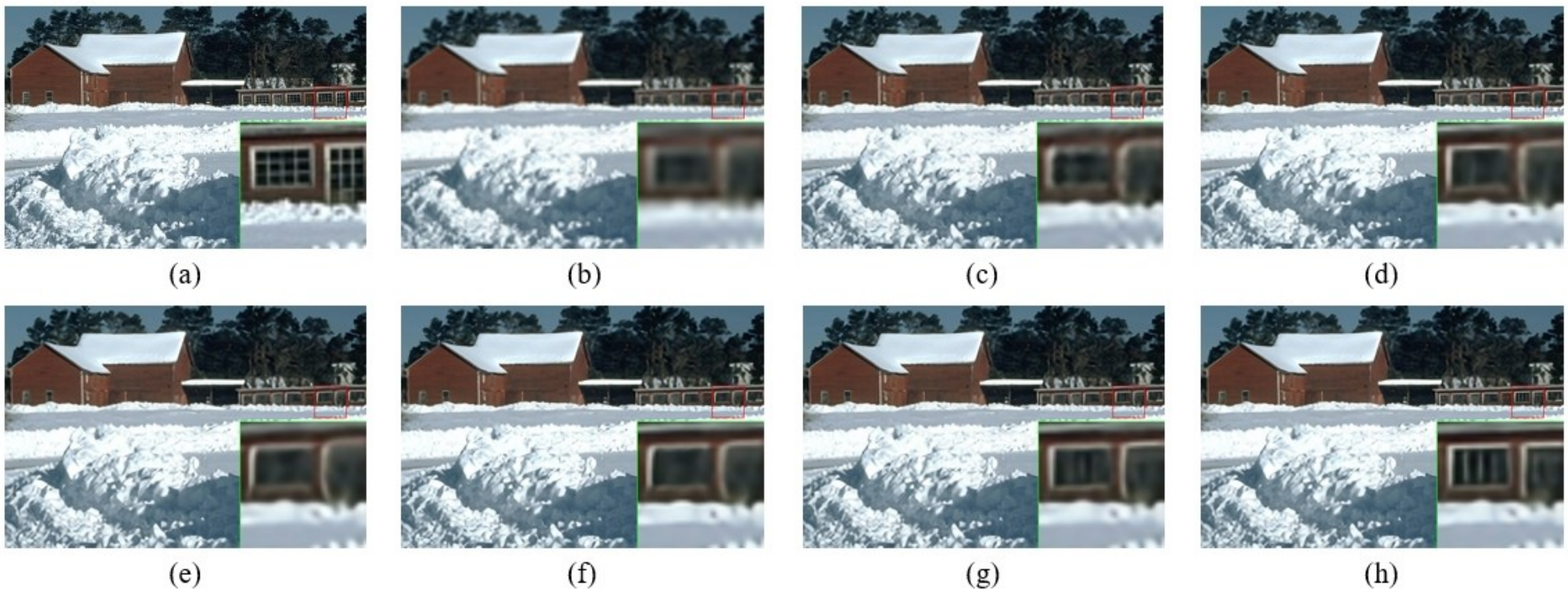

(a) HR 图像, (b)Bicubic, (c)SRCNN, (d)VDSR, (e)DRCN, (f)CARN-M, (g)LESRCNN 和 (h)CFSRCNN

图 11　不同图像超分辨方法在 BSD100 数据集上当倍数为 3 时的可视化效果.

**Fig.11 The visualization effects of different image super-resolution methods on the BSD100 dataset at multiples of 3**

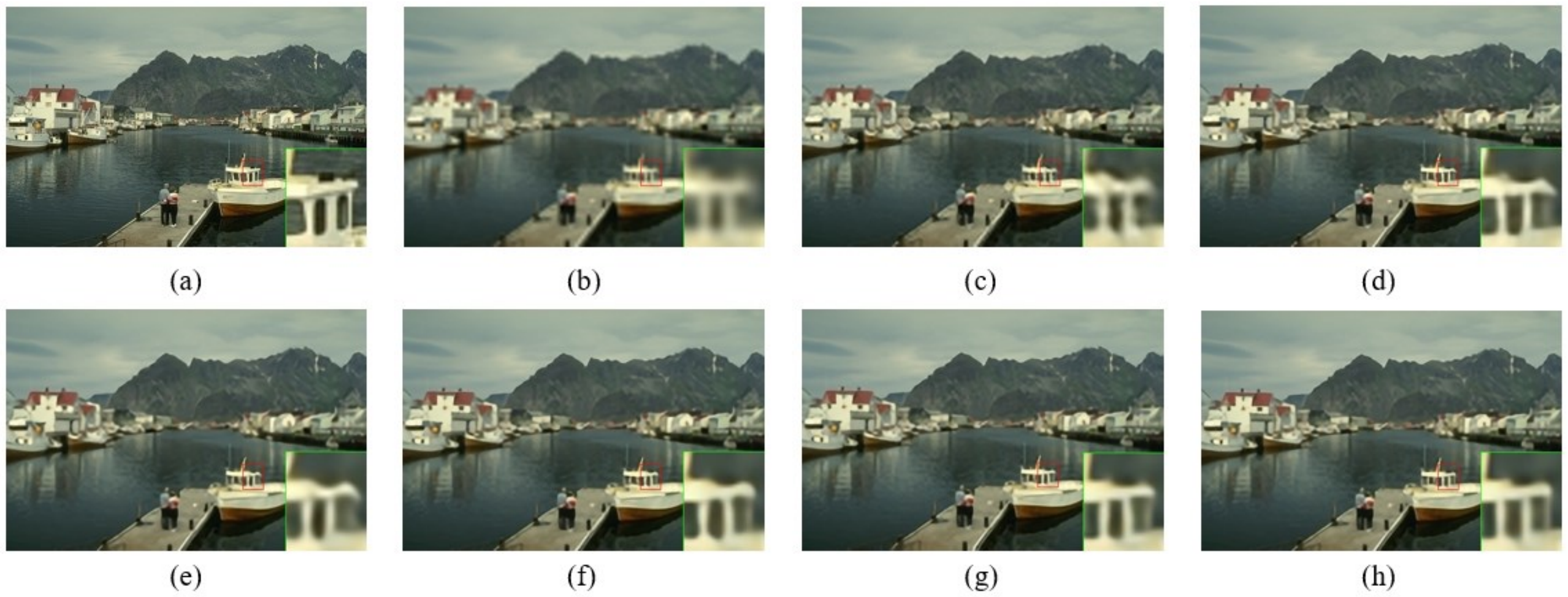

(a) HR 图像, (b)Bicubic, (c)SRCNN, (d)VDSR, (e)DRCN, (f)CARN-M, (g)LESRCNN 和 (h)CFSRCNN.

图 12 不同图像超分辨方法在 BSD100 数据集上当倍数为 4 时的可视化效果

**Fig.12 The visualization effects of different image super-resolution methods on the BSD100 dataset at multiples of 4**

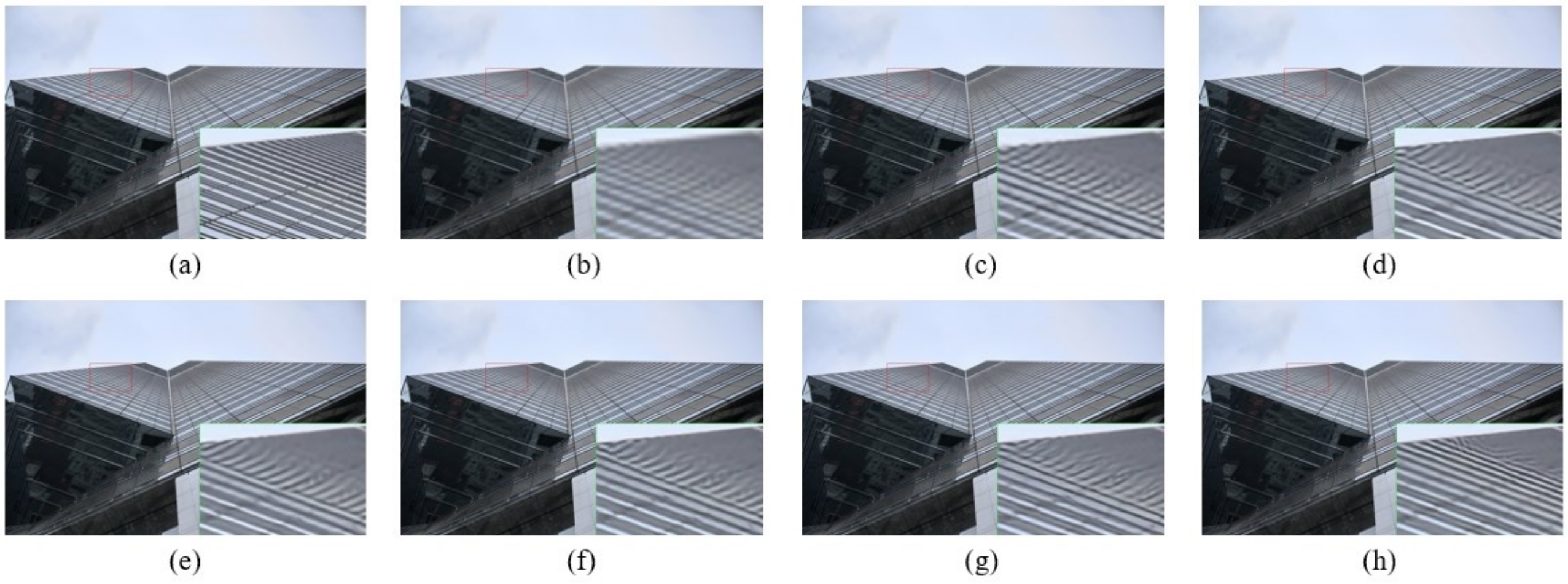

(a) HR 图像, (b)Bicubic, (c)SRCNN, (d)VDSR, (e)DRCN, (f)CARN-M, (g)LESRCNN 和 (h)CFSRCNN

图 13 不同图像超分辨方法在 Urban100 数据集上当倍数为 3 时的可视化效果

**Fig.13 The visualization effects of different image super-resolution methods on the Urban100 dataset at multiples of 3**

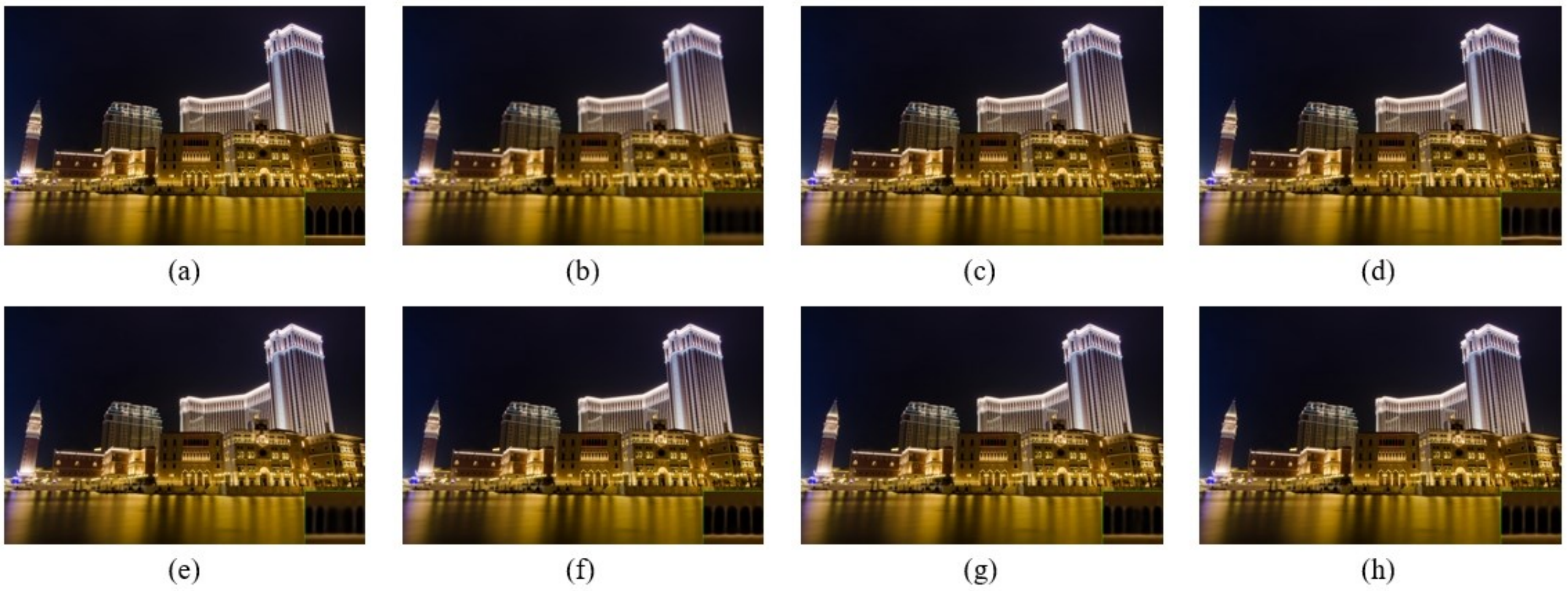

(a) HR 图像, (b)Bicubic, (c)SRCNN, (d)VDSR, (e)DRCN, (f)CARN-M, (g)LESRCNN 和 (h)CFSRCNN

图 14 不同图像超分辨方法在 Urban100 数据集上当倍数为 4 时的可视化效果.

**Fig.14 The visualization effects of different image super-resolution methods on the Urban100 dataset at multiples of 4**

## 4 潜在的研究点与挑战

本文介绍了图像超分辨技术的发展现状以及深度解析了基于插值的卷积神经网络图像超分辨方法和基于模块化的卷积神经网络，为读者深入理解图像超分辨方法提供参考。本节给出了深度学习在图像超分辨率领域的潜在方向和指出了一些尚未解决的问题。

### 4.1 潜在研究点

1）超分辨质量的衡量尺度：现有的超分辨方法主要采取像素级误差度量，例如两像素间距离或者两者组合。然而这些方法只封装了局部像素级信息，因此评价结果并不一定是感知上的可靠结果。例如，高 PSNR 和 SSIM 值的图像往往过于平滑，直观上的超分辨效果并不一定好。没有一种通用的感知度量尺度，可以在所有条件下很好地度量图像超分辨效果，因此，新的衡量尺度是一个重要的潜在研究点。

2）统一架构的图像超分辨模型：在真实的世界中，两个及两个以上的图像退化因素通常会同时出现来降低图像的质量。因此，如何能获得一个鲁棒性模型以解决复杂场景下的图像超分辨问题是一个重要的研究方向。

3）自监督图像超分辨：真实场景下的参考高清图像难以获取，使得已有的图像超分辨方法在真实世界中实用性减弱，因此，根据图像的属性以自监督方式开发出适合真实场景的图像超分辨性能是非常必要的。

4）大模型图像超分辨率：利用像素之间关联能促进更多的互补信息，提升图像超分辨性能。因此，大模型技术引导深度网络，能结合结构信息和像素信息，提高图像超分辨性能。

5）多模态图像超分辨率结合：由于复杂的拍摄场景、运动的拍摄设备以及运动的目标，导致单源的图像引导深度网络获得图像超分辨模型在真实场景中的应用受限。因此，多模态的图像超分辨方法是非常有必要的和研究前景的。

6）轻量级网络：移动设备的普及和算力受限的情况，使得高效、低功耗的图像处理算法需求增加。轻量级网络通过减少参数和计算量，实现高效的图像超分辨率，同时保持较高的图像质量。其优势在于能在低功耗设备上实时运行，适用于智能手机、无人机和边缘计算等应用场景。因此，轻量级网络的图像超分辨研究非常有必要的。

### 4.2 研究挑战

1）多种破坏因子的图像超分辨问题：在真实的世界中，由于相机抖动、运动的拍摄物体和相机以及复杂的拍摄背景等，会出现噪声、分辨率低等多种破坏因子，导致捕获的图像受损严重。因此，针对多种破坏因子的图像超分辨问题是有待解决的。

2）鲁棒的图像超分辨衡量指标：已有的图像超分辨方法通常采用PSNR和SSIM作为衡量指标，但获得可视化图像直观效果好，而对应方法的PSNR 和 SSIM 结果低。因此，制定鲁棒的图像超分辨衡量指标是迫在眉睫。

3）更大的超分辨放大倍率：目前，已有的超分辨模型基本无法解决超大的图像超分辨率问题，这限制了图像超分辨算法在如人群场景中人脸超分辨场景下的使用。因此，研究更大倍数的图像超分辨方法是迫在眉睫。

4）真实图像超分辨任务：已有的大部分图像超分辨方法都通过有参考的干净图像等有监督的方式获得图像超分辨模型，而在真实世界中，由于抖动的相机、运动物体以及拍摄的背景等导致已有图像超分辨模型不使用于真实的图像超分辨任务。因此，研究解决真实场景下图像超分辨方法是必要的。

5）无监督的图像超分辨：已有的无监督图像超分辨方法过分依赖于无标签的数据，如何获得高质量、多样化的数据是一个关键问题。在真实世界中，低分辨率图像的退化过程往往复杂多变，如何在无监督学习框架下有效地建模和处理这些复杂的退化过程，也是一个亟待解决的难点。

6）大模型的图像超分辨：大模型通常需要大量的计算资源和存储空间。此外，训练大模型还需要高质量和多样性的海量数据，获取和处理这些数据既耗时又复杂。其次，大模型的推理速度和能效需要优化，以便在实际应用中能够高效运行。大模型在图像超分辨率的应用仍然具有很大的挑战性。

7）硬件资源受限平台的图像超分辨率：移动设备和嵌入式系统等资源受限平台通常无法提供足够的计算资源和内存，而图像超分辨率算法通常需要大量的计算能力以及功耗。同时，如何在硬件资源有限的情况下保证处理速度和图像质量也是一个难点。在硬件资源受限平台应用图像超分辨仍面临严峻挑战。

## 5 结束语

本文比较和总结不同的卷积神经网络在图像超分辨上的应用，并给出了它们之间的联系与区别。首先，本文介绍了图像超分辨方法的发展状况。其次，本文根据图像属性和设备的要求，介绍了主流的图像超分辨卷积神经网络框架。随后，根据线性和非线性的缩放图像方式给出了基于插值的卷积神经网络图像超分辨方法（双三次插值算法、最近邻插值法、双线性插值算法）、基于模块化的卷积神经网络超分辨方法（转置卷积、亚像素层和元上采样模块），分析这些方法在非盲图像超分辨和盲图像超分辨问题上的动机、原理、区别和性能最后，本文给出卷积神经网络在图像超分辨的未来研究、挑战和总结全文。

## 参考文献:

**作者简介：**

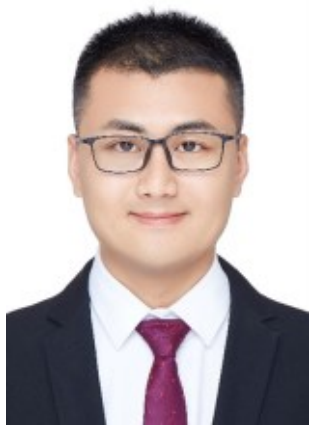

田春伟，教授，博士生导师，主要研究方向为图像复原和识别、图像生成。发表论文 80 余篇，6 篇 ESI 高被引论文，3 篇 ESI 热点论文、4 篇顶刊封面论文、1 篇国际模式识别会刊 Pattern Recognition 的 Best Paper Award、1 项中国图象图形学学会自然科学奖二等奖（排名第 1）。E-mail: chunweitian@hit.edu.cn。

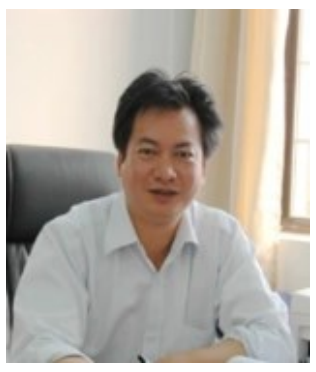

张师超，教授，博士生导师，主要研究方向为数据挖掘、机器学习。主持国家自然科学基金、国家 973 计划、国家 863 计划、澳大利亚 ARC（Australian Research Council）国家级项目共 12 项，及省部级项目 10 多项。在国际 SCI 核心学术期刊发表学术论文 200 多篇，发表国际会议论文 70 余篇。E-mail: zhangsc@mailbox.gxnu.edu.cn。

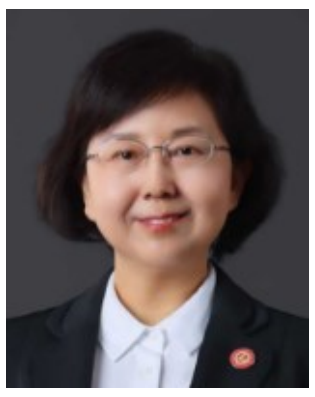

张艳宁，教授，博士生导师，主要研究方向为图像处理、模式识别、计算机视觉与智能信息处理。承担国防 973 项目、国家自然科学基金重点项目等国家级项目 40 余项，在国内外本领域权威期刊和重要国际会议上发表论文百余篇。

E-mail: ynzhang@nwpu.edu.cn。